\documentclass[10pt,twocolumn,letterpaper]{article}

\usepackage{iccv}
\usepackage{times}
\usepackage{epsfig}
\usepackage{graphicx}
\usepackage{amsmath}
\usepackage{amssymb}

\usepackage{bm}
\usepackage{booktabs}
\usepackage{siunitx}
\usepackage{stfloats}
\usepackage{multirow}
\usepackage{bbding} 
\usepackage{textcomp} 
\usepackage{pifont}
\usepackage{authblk}
\usepackage{subfigure}
\usepackage{makecell}
\usepackage[utf8]{inputenc}

\newcommand{\greencheck}{{\color{blue}\ding{51}}}

\usepackage[pagebackref=true,breaklinks=true,letterpaper=true,colorlinks,bookmarks=false]{hyperref}

\iccvfinalcopy 
\def\iccvPaperID{6834} 

\ificcvfinal\fi

\begin{document}

\title{A Hybrid Video Anomaly Detection Framework via Memory-Augmented Flow Reconstruction and Flow-Guided Frame Prediction}
\author[1]{Zhian Liu}
\author[1\thanks{Corresponding author: nieyongwei@scut.edu.cn}]{Yongwei Nie}
\author[2]{Chengjiang Long}
\author[3]{Qing Zhang}
\author[1]{Guiqing Li}
\affil[1]{School of Computer Science and Engineering, South 
China University of Technology, China}
\affil[2]{JD Finance America Corporation, Mountain View, CA, USA}
\affil[3]{School of Computer Science and Engineering, Sun Yat-sen University, China}

\renewcommand\Authands{ and }

\maketitle
\ificcvfinal\thispagestyle{empty}\fi

\begin{abstract}
In this paper, we propose $\text{HF}^2$-VAD, a Hybrid framework that integrates Flow reconstruction and Frame prediction seamlessly to handle Video Anomaly Detection. Firstly, we design the network of ML-MemAE-SC (Multi-Level Memory modules in an Autoencoder with Skip Connections) to memorize normal patterns for optical flow reconstruction so that abnormal events can be sensitively identified with larger flow reconstruction errors. More importantly, conditioned on the reconstructed flows, we then employ a Conditional Variational Autoencoder (CVAE), which captures the high correlation between video frame and optical flow, to predict the next frame given several previous frames. By CVAE, the quality of flow reconstruction essentially influences that of frame prediction. Therefore, poorly reconstructed optical flows of abnormal events further deteriorate the quality of the final predicted future frame, making the anomalies more detectable. Experimental results demonstrate the effectiveness of the proposed method. Code is available at \href{https://github.com/LiUzHiAn/hf2vad}{https://github.com/LiUzHiAn/hf2vad}.

\end{abstract}

\vspace{-0.7cm}
\section{Introduction}
\begin{figure}
\centering
\includegraphics[width=0.48\textwidth]{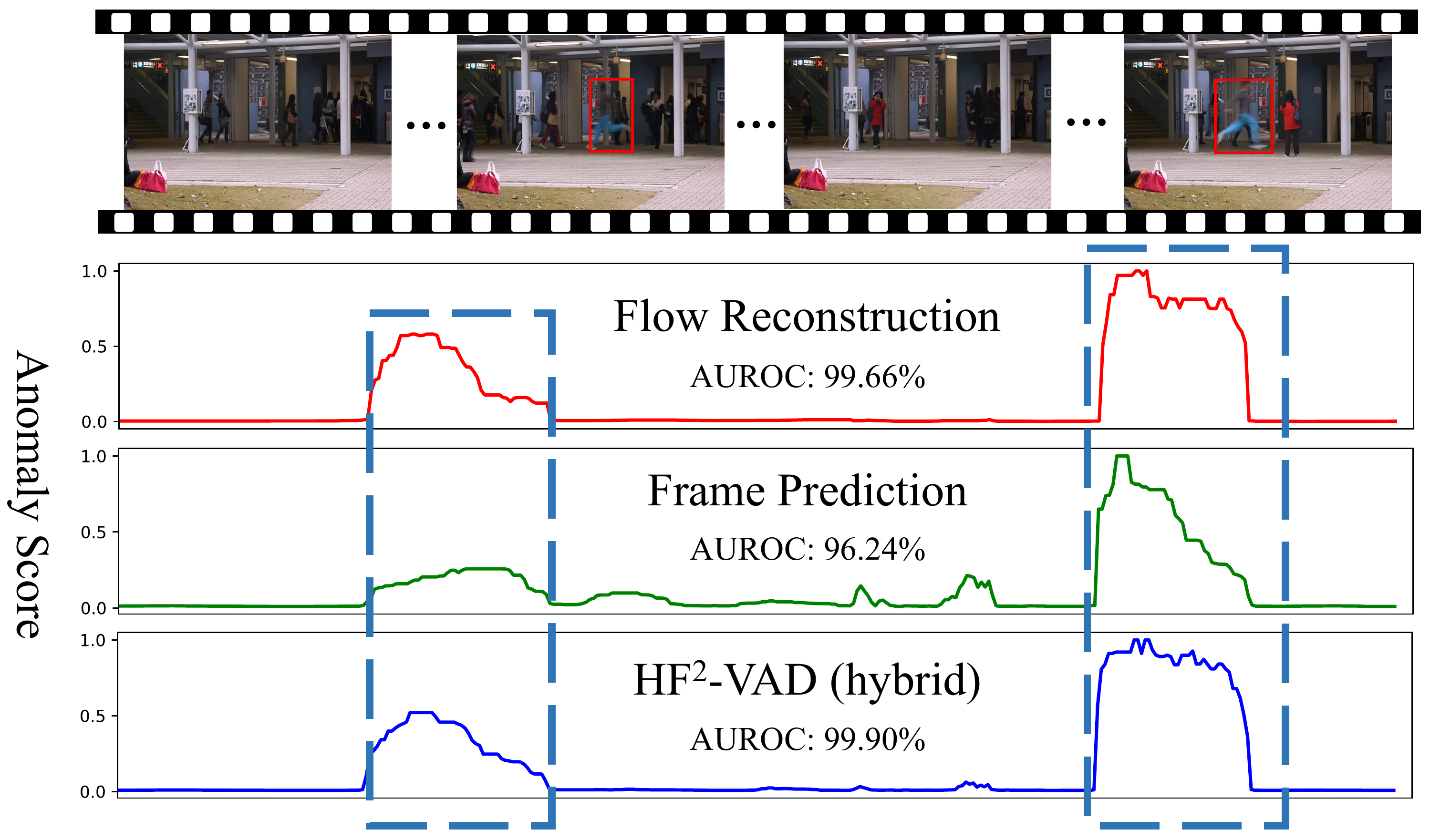}
\caption{An anomaly example from CUHK Avenue~\cite{lu2013abnormal} dataset. Here ``{\em running}" is identified as an anomaly event on the walking avenue, and the blue dashed rectangles denote ground-truth abnormal sections where a person is running. The blue curve is the result of HF$^2$-VAD, the red curve is the result with flow reconstruction only, and the green curve is the result with a CVAE based frame prediction model conditioned on original flows. Area Under the Receiver Operation Characteristic (AUROC) is calculated. As can be seen, HF$^2$-VAD that combines flow reconstruction and reconstructed-flow guided future prediction performs the best.}
\label{fig:introduction_example}
\vspace{-0.6cm}
\end{figure}

Video Anomaly Detection (VAD) refers to the identification of events that do not conform to expected behaviors~\cite{chandola2009anomaly} in a video, with one example shown in Figure~\ref{fig:introduction_example}. This is an open and very challenging task as abnormal events usually much less happen than normal ones and the forms of abnormal events are unbounded in practical applications~\cite{liu2018future}. Obviously, it is impossible to collect all kinds of abnormal data in advance.
Therefore, a typical solution to video anomaly detection is to train an unsupervised learning model on normal data, and those events or activities that are recognized by the trained model as outliers are then deemed as anomalies.

Nowadays deep learning has shown great success in many real-world tasks such as visual recognition~\cite{Hu:TIP2021, Hu:arXiv2021, Hua:ICCV2013B, Long:ICCV2015}, object detection~\cite{Islam:AAAI2021,Long:IJCV2016, Long:CVPR2017, Hua:TPAMI2018}, shadow detection and removal~\cite{Ding:ICCV2019, Wei:CGF2019, Zhang:AAAI2020, Zhang:CGF2020,Islam:CVPR2020}, trajectory prediction~\cite{Shi:CVPR2021}, and image captioning~\cite{Dong:MM2021}. Rather than traditional hand-crafted feature based methods~\cite{adam2008robust,kim2009observe,benezeth2009abnormal,mahadevan2010anomaly}, a lot of
modern deep neural network based methods~\cite{hasan2016learning,xu2017detecting,zhao2017spatio,luo2017remembering,liu2018future, lu2019future, gong2019memorizing, nguyen2019anomaly,park2020learning, yu2020cloze,tian2021weakly,li2020decoupled} have been proposed for VAD. In the era of deep learning, reconstruction and future prediction are two prevalent VAD paradigms. Reconstruction-based methods~\cite{hasan2016learning,luo2017remembering,gong2019memorizing,nguyen2019anomaly,fan2020video,park2020learning} typically train autoencoders on normal data. At test time, abnormal data often incurs larger reconstruction errors, making them detectable from normal ones. Taking advantage of temporal characteristics of video frames, prediction-based methods~\cite{liu2018future,lu2019future,yu2020cloze} train a network to predict the next frame based on the given previous frames and use prediction error for anomaly measuring. Recently, several works~\cite{zhao2017spatio,ye2019anopcn,morais2019learning} are proposed to combine these two paradigms in a hybrid manner. Although these state-of-the-art methods have been able to detect anomalies in most cases, the results are still far from perfect.

\begin{figure*}[ht] 
\vspace{-0.6cm}
\centering
\includegraphics[width=0.98\textwidth]{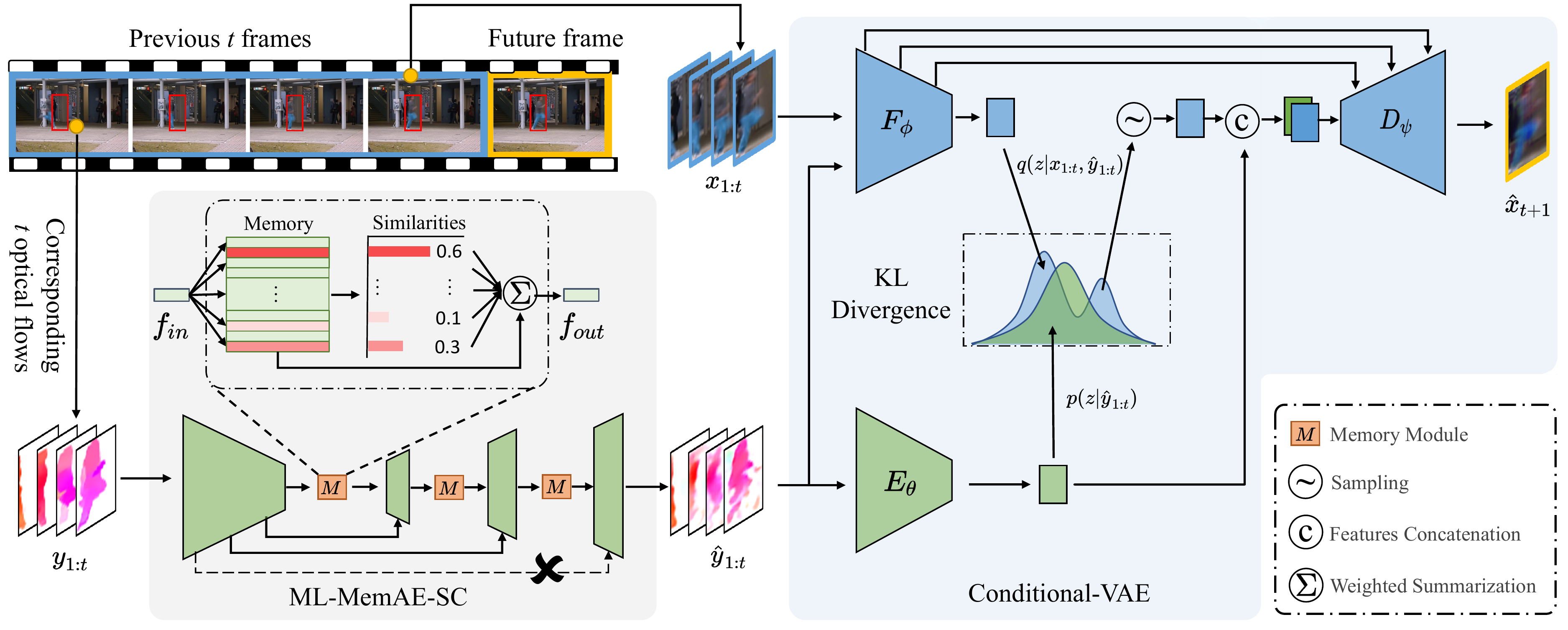}
\caption{Overview of the proposed HF$^2$-VAD which integrates flow reconstruction and frame prediction into a unified framework. We first reconstruct the optical flows $y_{1:t}$ by an autoencoder with multi-level memory modules and skip connections to obtain $\hat{y}_{1:t}$. Then, the reconstructed optical flows together with the video frames $x_{1:t}$ are used by a CVAE model to predict the next future frame. If an abnormal event occurs, (1) the reconstructed optical flows $\hat{y}_{1:t}$ will show significant reconstruction error to its ground truth $y_{1:t}$, (2) using the $\hat{y}_{1:t}$ as a condition to guide future frame prediction, the prediction error (\ie, difference between $\hat{x}_{t+1}$ and $x_{t+1}$) will be enlarged further.} 
\label{fig:final}
\vspace{-0.3cm}
\end{figure*}

In this paper, we propose a novel {\em hybrid} framework in a combination of {\em flow} reconstruction and flow-guided {\em frame} prediction, named as ``HF$^2$-VAD", for {\em video anomaly detection}. As illustrated in Figure~\ref{fig:final}, the Conditional VAE (CVAE) based future frame prediction model accepts both previous video frames and optical flows as input. But instead of original flows, we reconstruct them in advance and then input the reconstructed flows into the CVAE model.

Inspired by~\cite{gong2019memorizing,park2020learning}, we design a Multi-Level Memory-augmented Autoencoder with Skip Connections (ML-MemAE-SC) for optical flow reconstruction. Multiple memory modules are employed to memorize normal patterns at different feature levels, while skip connections are added between encoder and decoder to compensate for the strong information compression due to the memories. We observe that such a well-designed flow reconstruction network can reconstruct normal flows more clearly while producing larger reconstruction error for abnormal input.

We use the model of CVAE~\cite{NIPS2015_8d55a249} for future frame prediction. On one hand, it takes the reconstructed flows by ML-MemAE-SC as condition, unifying the reconstruction module into the prediction pipeline naturally. On the other hand, by maximizing the quantity of evidence lower bound (ELBO) induced from the variables of observed video frames and reconstructed optical flows, the CVAE module essentially encodes the consistency between the input frames and flows when taking them for the future frame prediction.

The above design facilitates to utilize the quality gap between the reconstructed normal and abnormal flows to improve the VAD accuracy of the CVAE-based prediction module. That is, the reconstructed normal flows usually have higher quality, with which the prediction module can successfully predict the future frame with smaller prediction error. In contrast, the reconstructed abnormal flows usually have lower quality, thus leading to future frame with larger prediction error. We use both the flow reconstruction and frame prediction errors as our final anomaly detection cues. In summary, the following aspects distinguish our work from the previous works~\cite{liu2018future, gong2019memorizing, fan2020video,yu2020cloze,nguyen2019anomaly,lu2019future,yu2020cloze}:
\begin{itemize}
    \item First of all, multi-level memory modules are utilized in an encoder-decoder structure with skip connections, which guarantees normal patterns are well memorized so that the abnormal events or activities are sensitively identified.
    \item Second, we design the HF$^2$-VAD hybrid method that predicts future frame from both previous video frames and the corresponding optical flows but with the flows being reconstructed beforehand. The reconstruction error enlarges the prediction error so that anomaly can be more easily detected.
    \item Finally, we conduct extensive experiments on three public datasets which show that our proposed HF$^2$-VAD achieves better anomaly detection performance than state-of-the-art methods.
\end{itemize}

\vspace{-0.4cm}
\section{Related work}
\textbf{Video anomaly detection.} 
Feature extraction followed by a normality modeling was a popular paradigm in early works~\cite{piciarelli2008trajectory, adam2008robust, kim2009observe}. With the powerful feature representation capability of CNNs, many deep learning based VAD methods have been proposed in recent years. Reconstruction-based and prediction-based methods are the two mainstreams in the VAD community. In the reconstruction camp, autoencoders (AEs) are widely leveraged to reconstruct the training data, such as Conv-AE~\cite{hasan2016learning}, ConvLSTM-AE~\cite{luo2017remembering} and GMFC-VAE~\cite{fan2020video}. They assume the anomalies are hard to be reconstructed well by the AE trained only on normal data. But this assumption not always holds since AE sometimes generalizes so well~\cite{gong2019memorizing}. In the prediction camp, Liu \etal~\cite{liu2018future} first propose to predict the future frame and use prediction error as the anomaly indicator. Lu \etal~\cite{lu2019future} then introduce a variational Conv-LSTM for prediction. Yu \etal~\cite{yu2020cloze} borrow the cloze test concept in language study and apply it to video event completion, in which multiple models are built to predict each frame or flow in a clip separately. However, these methods simply take previous frames as input to predict the future, ignoring the correlation between optical flow and video frame. Further, some works joint these two paradigms together to develop hybrid approaches. For example, an AE comprising a shared encoder and two separate decoders is proposed in~\cite{nguyen2019anomaly}, to reconstruct frame and predict flow separately. Ye \etal~\cite{ye2019anopcn} decompose reconstruction into prediction and refinement, proposing a predictive coding network. In our work, we integrate flow reconstruction and frame prediction seamlessly, making the quality of flow reconstruction essentially influence that of frame prediction.

\textbf{Memory networks.} Memory module in neural networks has attracted much attention recently. Graves \etal~\cite{graves2016hybrid} introduce a differentiable neural computer model which consists of a neural network to extract features and an external memory module to store information explicitly. In order to suppress the generalization capability of AE, Gong \etal~\cite{gong2019memorizing} propose a Memory-augmented Autoencoder (MemAE) for anomaly detection. MemAE receives information from the encoder and then uses it as a query to retrieve some similar memory slots which are then combined to yield new encoding features for the decoder to reconstruct. The MemAE is trained on normal data, thus encouraged to store normal patterns in the memory. Park \etal ~\cite{park2020learning} follow this trend and present a more compact memory that can be updated during testing. In their works, the memory module is only placed at the bottleneck. We extend it and propose a multi-level memory-augmented autoencoder with skip connections which captures the normal patterns at different feature levels and train it on optical flows.

\textbf{VAE and CVAE.} Along with the advances in discriminative models, generative models have also made great progress, such as GAN \cite{goodfellow2014generative} and VAE \cite{kingma2013auto}. In particular, VAE is a directed graphical model with latent variables, which includes a recognition process and a generative process. In the generative process (\ie, decoding), the data $x$ is generated by the distribution $p_\theta(x|z)$ when given latent variables $z$. Kingma and Welling~\cite{kingma2013auto} then introduce a recognition process $q_\phi(z|x)$ to approximate the intractable true posterior. Both the recognition and generative distributions can be learned by maximizing the log-likelihood of the $p(x)$ through the surrogate evidence lower bound (ELBO) objective function. To address the structured prediction problem, Sohn \etal~\cite{NIPS2015_8d55a249} model the distribution of the output space conditioning on the input observation and propose CVAE. Let $x$, $y$ and $z$ represent the output data, observed condition and latent variables, CVAE is composed of a recognition network $q_\phi(z|x,y)$, a conditional prior network $p_\theta(z|y)$ and a generation network $p_\theta(x|y,z)$. Esser~\etal \cite{esser2018variational} follow the CVAE framework and design a variational UNet that can disentangle the appearance and shape information of an image by which the image generating process can be well controlled. Different from them, we propose to predict future frame by CVAE with optical flow as the condition.

\section{Methodology}
As illustrated in Figure~\ref{fig:final}, our framework HF$^2$-VAD is composed of two components: Multi-Level Memory-augmented Autoencoder with Skip Connections (ML-MemAE-SC) for flow reconstruction followed by Conditional Variational Autoencoder (CVAE) for frame prediction. The whole framework is trained on normal data. At test time, both the reconstruction and prediction errors, \ie, the difference between $y_{1:t}$ and $\hat{y}_{1:t}$, and $x_{t+1}$ and $\hat{x}_{t+1}$, are used for anomaly detection.

In the following sections, we introduce ML-MemAE-SC first, and then the CVAE based future frame prediction model. Finally, we show how to use our model for anomaly detection.

\begin{figure*}[!ht] 
\vspace{-0.8cm}
\centering
 \subfigure[MemAE]{\includegraphics[width=0.22\textwidth]{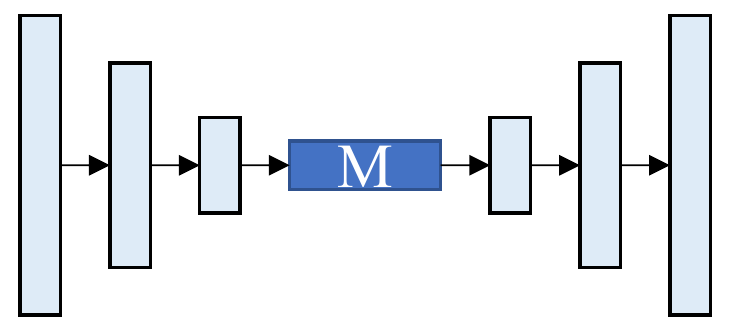}}
 \hspace{0.5cm}
 \subfigure[ML-MemAE]{\includegraphics[width=0.33\textwidth]{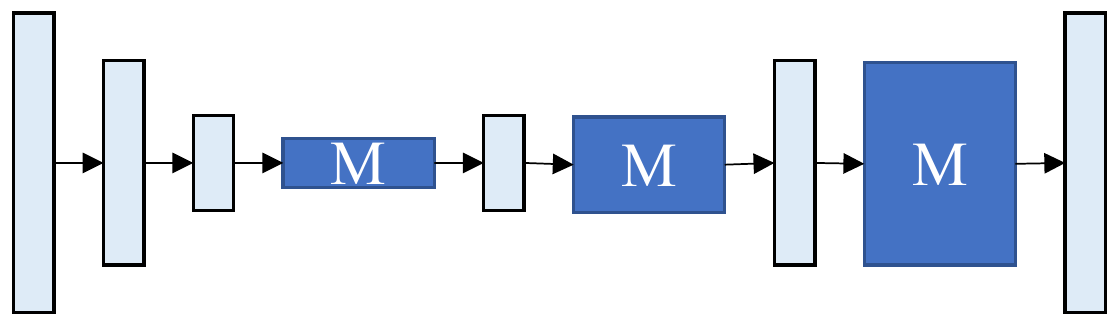}}
 \hspace{0.5cm}
 \subfigure[ML-MemAE-SC]{\includegraphics[width=0.34\textwidth]{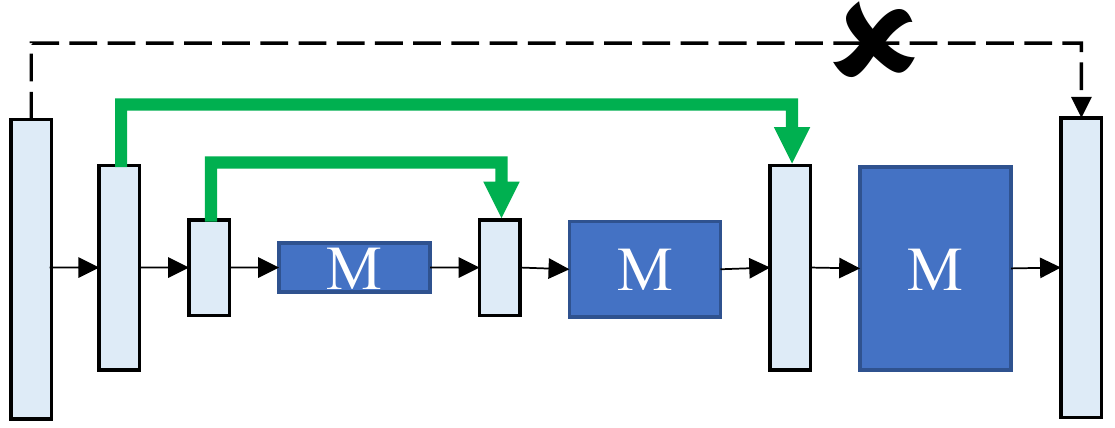}}\\
\caption{(a) Structure of MemAE in which a memory module is placed at the bottleneck. (b) Extending MemAE with more memory modules in other levels of the decoder. (c) On the basis of (b), skip connections are added, but the black dotted skip connection should not be added, otherwise the functions of all the memory modules would be overridden.} 
\label{fig:mlmaunet}
\vspace{-0.4cm}
\end{figure*}
\subsection{Multi-Level Memory-augmented Autoencoder with Skip Connections}\label{sec:mlmemunet}
Placing a memory module at the bottleneck of AE is a recent development in VAD community~\cite{gong2019memorizing, park2020learning}. Figure~\ref{fig:mlmaunet} (a) shows this kind of Memory-augmented Autoencoder (MemAE). However, we observe that using just one memory can not guarantee all the normal patterns to be remembered, and abnormal input still has a certain chance to be reconstructed well. A natural extension to MemAE is placing more memory modules in other levels of AE as shown in Figure~\ref{fig:mlmaunet} (b), but too many memories lead to excessive information filtering, degrading the network to remember the most representative normal patterns rather than all needed ones. We solve this problem by adding skip connections between encoder and decoder, obtaining the Multi-Level Memory-augmented Autoencoder with Skip Connections (ML-MemAE-SC), as shown in Figure~\ref{fig:mlmaunet} (c). On one hand, the skip connections directly transfer encoding information to the decoder, providing more information for memories in different levels to discover normal patterns. On the other hand, with higher-level encoding features, though being filtered by the memories, the network can decode the input more easily. At test time, the proposed ML-MemAE-SC can reconstruct normal data clearly while performs poorly for abnormal data. To make it easy for readers to validate this, we conduct a toy example to explore many memory-augmented autoencoder variants and demonstrate the efficacy of the proposed ML-MemAE-SC. See Figure~\ref{fig:ml_ablation}.

It is worth noting that the outermost skip connection, \ie, the black dotted one in Figure~\ref{fig:mlmaunet} (c), should not be added. Otherwise the reconstruction might be fulfilled by the highest-level encoding-decoding information, making all other lower-level encoding, decoding and memory blocks not work.  

We design a four-level ML-MemAE-SC, including three encoding-decoding levels and the bottleneck. In each level of the encoder, we stack two convolution blocks followed by a downsampling layer. In each level of the decoder, we first copy the feature map from the encoder and then concatenate it with the upsampled feature maps of the lower level. The concatenation then sequentially passes through two convolution blocks, a memory module, and an upsampling layer. In our implementation, a convolution block contains three layers: a convolution layer, a batch-normalization layer~\cite{ioffe2015batch} and a ReLU activation layer~\cite{maas2013rectifier}. The downsampling and upsampling layers are implemented by convolution and deconvolution~\cite{noh2015learning}.

For the memory modules, we adopt a similar implementation of~\cite{gong2019memorizing}. Each memory module is actually a matrix $\mathbf{M} \in \mathbb{R}^{N \times C}$. Each row of the matrix is called a slot $\mathbf{m}_{i}$ with $i=1,2,3,...,N$. The role of a memory module is to represent the features fed into it by the weighted sum of similar memory slots, thus has the capability of remembering normal patterns when trained on normal data.

To train ML-MemAE-SC, we can feed normal video, image or optical flow into it, and try to reconstruct the input data. Let $y$ be the input data, and $\hat{y}$ be the reconstructed result, we minimize the $\ell_2$ distance between $y$ and $\hat{y}$ as the reconstruction loss:
\vspace{-0.15cm}
\begin{equation}
\mathcal{L}_{recon}=||y-\hat{y}||_2^2.
\vspace{-0.15cm}
\end{equation}
Following \cite{gong2019memorizing}, we add the entropy loss on the matching probabilities $\hat{w}_{i}$ for each memory module as:
\vspace{-0.15cm}
\begin{equation}
\mathcal{L}_{ent}=\sum_{i=1}^{M}  \sum_{k=1}^{N}-\hat{w}_{i, k} \log \left(\hat{w}_{i, k}\right),
\vspace{-0.25cm}
\end{equation}
where $M$ is the number of memory modules and $\hat{w}_{i, k}$ is the matching probabilities for the $k$-th slot in the $i$-th memory module. We balance the above two loss functions to obtain the following loss function to train ML-MemAE-SC:
\vspace{-0.2cm}
\begin{equation}
\mathcal{L}_{ML-MemAE-SC}=\lambda_{recon}\mathcal{L}_{recon}+\lambda_{ent}\mathcal{L}_{ent}.
\vspace{-0.1cm}
\end{equation}

\subsection{Conditional Variational Autoencoder for Future Frame Prediction}\label{sec:MPFN}
\vspace{-0.1cm}
Future frame prediction is another prevalent VAD paradigm, often obtaining better anomaly detection accuracy than reconstruction-based methods~\cite{yu2020cloze,park2020learning}. Future frame prediction tries to model $p(x_{t+1}|x_{1:t})$ such that the next frame $x_{t+1}$ can be generated given $x_{1:t}$. Many works have explored the usage of optical flow as the auxiliary information to increase prediction accuracy~\cite{liu2018future,nguyen2019anomaly}, but to our best knowledge there is no work that directly models $p(x_{t+1}|x_{1:t},y_{1:t})$, where $y_{1:t}$ represents previous $t$ flows.

Note that $x_t$ can be warped to $x_{t+1}$ given optical flow $y_t$, thus a vanilla network that directly maps $x_{1:t}$ and $y_{1:t}$ to $x_{t+1}$  may learn a trial mapping. We observe that $x_{1:t}$ and $x_{t+1}$ are from a very short duration in a video, and they are very similar to each other in content. It is reasonable to assume that $x_{1:t}$ and $x_{t+1}$ are determined by the same hidden variables $z$ that control the content information. We thus resort to Conditional Variational Autoencoder (CVAE) as the generative model for modeling $p(x_{t+1}|x_{1:t},y_{1:t})$, in which we compute $z$ from $x_{1:t}$, and then generate $x_{t+1}$ from $z$, with $y_{1:t}$ as the conditions.

Formally, we have the following ELBO :
\begin{equation}
\vspace{-0.1cm}
log\ p(x_{t+1}|y_{1:t})\geq \mathbb{E}_{q} \log \frac{p(x_{t+1}|z, y_{1:t}) p(z|y_{1:t})}{q(z| x_{t+1}, y_{1:t})}. \label{eq:cvae}
\end{equation}
Replacing $q(z|x_{t+1}, y_{1:t})$ in Eq.~\ref{eq:cvae} by $q(z|x_{1:t}, y_{1:t})$ yields:
\begin{align}
\vspace{-0.1cm}
log\ p(x_{t+1}|y_{1:t})&\geq \mathbb{E}_{q} \log \frac{p(x_{t+1}|z,y_{1:t}) p(z|y_{1:t})}{q(z|x_{t+1}, y_{1:t})} \nonumber \\
&\approx \mathbb{E}_{q} \log \frac{p(x_{t+1}|z,y_{1:t}) p(z|y_{1:t})}{q(z|x_{1:t} ,y_{1:t})} \nonumber \\
&=-KL[q(z|x_{1:t},y_{1:t})\ || \ p(z|y_{1:t})] \nonumber \\ 
&\quad\quad +\mathbb{E}_{q}[\log p(x_{t+1}|z,y_{1:t})], \label{eq:cvae_our}
\end{align}
where $KL$ denotes Kullback-Leibler divergence.

Guided by Eq.~\ref{eq:cvae_our}, we design our future prediction model as shown in Figure~\ref{fig:final}. We have two encoders $E_{\theta}$ and $F_{\phi}$, and one decoder $D_{\psi}$. $E_{\theta}$ encodes optical flows $y_{1:t}$ to obtain $E_{\theta}(y_{1:t})$ from which the prior distribution $p(z|y_{1:t})$ can be obtained. $F_{\phi}$ admits the concatenation of $x_{1:t}$ and $y_{1:t}$ and outputs features $F_{\phi}(x_{1:t},y_{1:t})$ from which the posterior distribution $q(z|x_{1:t},y_{1:t})$ can be obtained. During training, we sample $z$ from the posterior distribution, and concatenate $z$ with the conditions $E_{\theta}(y_{1:t})$, which are finally sent to the decoder $D_{\psi}$ to generate the future frame $\hat{x}_{t+1}$. Inspired by the Variational UNet proposed in~\cite{esser2018variational}, we add skip connections between $F_{\phi}$ and $D_{\psi}$ to help generating $x_{t+1}$.

We assume $p(x_{t+1}|z,y_{1:t})$, $p(z|y_{1:t})$ and $q(z|x_{1:t},y_{1:t})$ in Eq.~\ref{eq:cvae_our} are all parametric Gaussian distributions. Hence, as the common practice in VAE~\cite{kingma2013auto}, we arrive at the following loss function containing two parts:
\begin{align}
\vspace{-0.05cm}
\mathcal{L}_{CVAE}=&KL[q(z|x_{1:t},y_{1:t})||p(z|y_{1:t})] \nonumber \\ 
& +||x_{t+1}-\hat{x}_{t+1}||_2^2,\label{eq:11}
\end{align}
where $x_{t+1}$ is the ground truth future frame. 
Following~\cite{liu2018future}, we also define a gradient loss:
\begin{align}
\vspace{-0.05cm}
\mathcal{L}_{gd}(X,\hat{X})=&\sum_{i, j}\Big|| X_{i, j}-X_{i-1, j}|-| \hat{X}_{i, j}-\hat{X}_{i-1, j}|\Big|  \nonumber  \\ 
 &\Big|| X_{i, j}-X_{i, j-1}|-| \hat{X}_{i, j}-\hat{X}_{i, j-1}|\Big|, \label{eq:12} 
\end{align} 
where $i,j$ indicate spatial pixel position in an image. Combining Eq.~\ref{eq:11} and the gradient loss between the predicted future frame and its ground truth, we train our CVAE model by the following loss function:
\begin{equation}
\vspace{-0.05cm}
\mathcal{L}=\lambda_{CVAE}L_{CVAE}+\lambda_{gd}\mathcal{L}_{gd}(\hat{x}_{t+1},x_{t+1}),
\end{equation}
where $\lambda_{CVAE}$ and $\lambda_{gd}$ are the balancing hyper-parameters.

\textbf{Deterministic \vs stochastic future prediction during testing}. At test time, we can sample $z$ stochastically in order to generate the future frame. But this would synthesize slightly different future frames at different time. In order to predict the future frame deterministically, we use the mean of the posterior distribution $q(z|x_{1:t},y_{1:t})$ as the sampled $z$ at test time, and all the experimental results of our method in this paper are obtained under this sampling strategy. But note that this two kinds of sampling strategies have similar anomaly detection performance. Please see Appendix~\ref{appendix:sampling} for more details.

\subsection{Anomaly detection}
At test time, our anomaly score is composed of two parts: (1) the flow reconstruction error as $S_r=||\hat{y}_{1:t}-y_{1:t}||_2^2$ and (2) the future frame prediction error as $S_p= ||\hat{x}_{t+1}-x_{t+1}||_2^2$. We obtain the anomaly score by fusing the two errors using a weighted sum strategy as:
\begin{equation}
S=w_r\cdot \frac{S_r-\mu_r}{\sigma_r}+w_p\cdot \frac{S_p-\mu_p}{\sigma_p},\label{eq:ensemble}
\end{equation}
where $\mu_r,\sigma_r,\mu_p,\sigma_p$ are means and standard deviations of reconstruction errors and prediction errors of all the training samples, $w_r$ and $w_p$ are the weights of the two scores.

\section{Experiments} \label{sec:experiments}

\subsection{Toy experiments on MNIST}\label{sec:toy_exp}

\begin{figure}[t]
\centering
\vspace{-0.5cm}  
\includegraphics[width=0.45\textwidth]{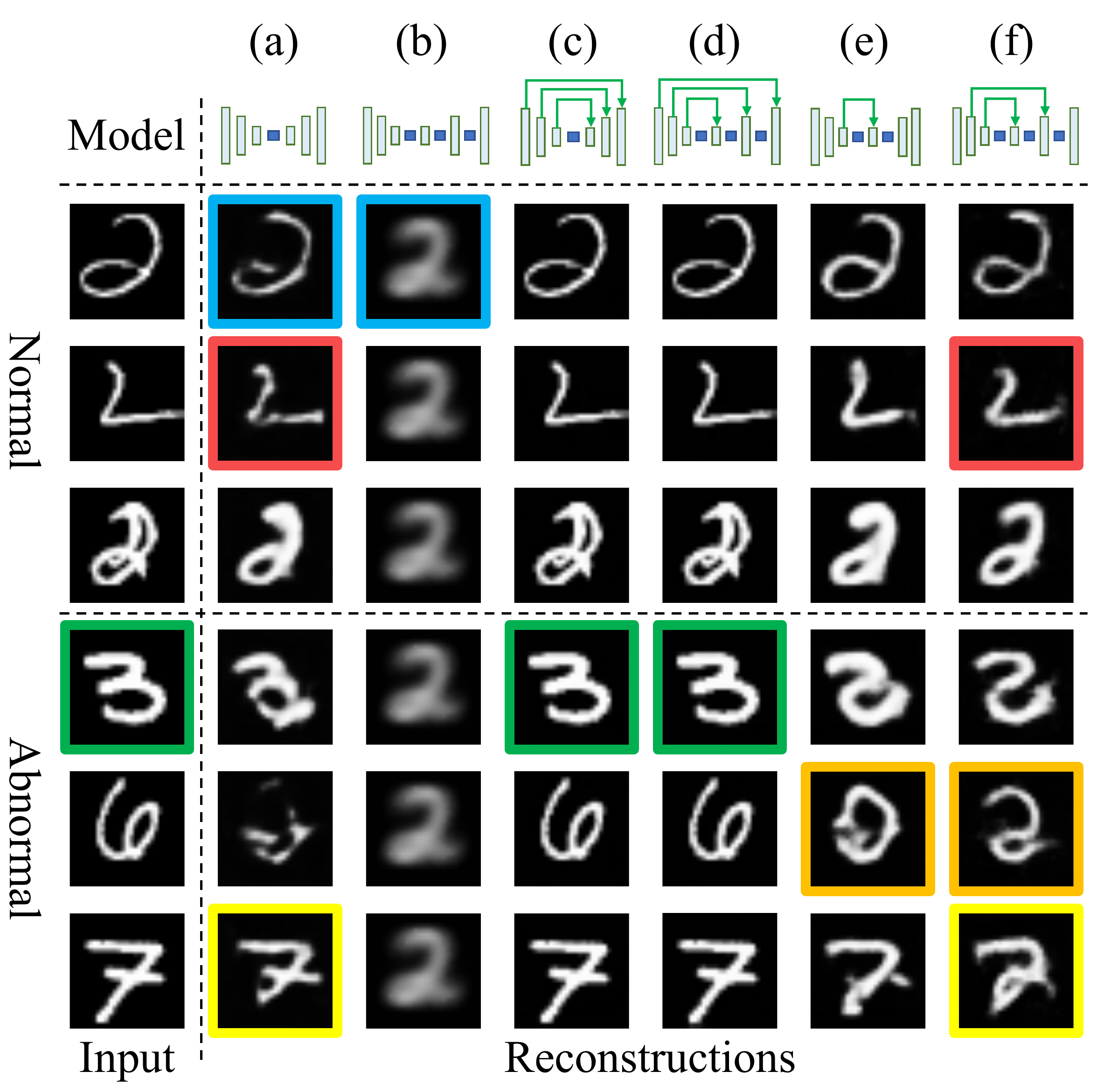}
\caption{Visualization examples of memory-augmented autoencoder variants for reconstruction task on MNIST~\cite{lecun2010mnist} dataset. }
\label{fig:ml_ablation}
\vspace{-0.3cm}
\end{figure}

We conduct toy anomaly detection experiments on the MNIST~\cite{lecun2010mnist} dataset to explore the variations of memory-augmented autoencoders. We use ``2'' of the training set of MNIST as the normal data to train our models, and use the testing set of MNIST as our test data in which digits other than ``2'' are abnormal. We compare among 6 variants, the architectures of which are shown in the top row of Figure~\ref{fig:ml_ablation} from (a) to (f). (a) is the original MemAE. (b) is a variant of (a) by adding memory modules at all other levels of the decoder. (c) is another variant of (a) by adding skip connections between encoder and decoder at all levels. (d) is the combination of (b) and (c). In (e), we add one more memory to MemAE in the next level after bottleneck and also a skip connection for that level. In (f), we extend (e) by adding another pair of memory and skip connection in the next level further, which is the proposed ML-MemAE-SC.

As shown in Figure~\ref{fig:ml_ablation}, MemAE reconstructs normal ``2'' very well, but it also reconstructs ``3'' and ``7'' successfully which however are anomalies. By simply adding more memory modules in (b), all tested inputs are reconstructed to a blurry ``2'', demonstrating that cascaded memories are too aggressive to filter out all useful information. The models in (c) and (d) reconstruct both normal and abnormal inputs very clearly, due to the reason that the function of the memories is blocked by the outer skip connections. In (e) and (f), skip connection and memory are added in pairs, with skip connection providing additional features for the corresponding memory to learn more normal patterns. As can be seen, the reconstructed outputs of normal data in (e) and (f) are more clear than those in (a), while for the abnormal inputs other than ``2'', the models in (e) and (f) try to reconstruct them to ``2'' which cannot be accomplished by MemAE, indicating normality is better learned in (e) and (f). Since (f) has one more pair of skip connection and memory, the results of (f) are even better than those of (e). 

\subsection{Video Anomaly Detection Experiments}
\textbf{Datasets.} To evaluate both qualitative and quantitative results of the proposed HF$^2$-VAD and compare it with state-of-the-art algorithms, we conduct experiments on three public video anomaly detection datasets, {\em i.e.}, UCSD Ped2 \cite{mahadevan2010anomaly}, CUHK Avenue \cite{lu2013abnormal} and ShanghaiTech~\cite{luo2017revisit}. {(a) UCSD Ped2} consists of 16 training videos and 12 testing videos, acquired with a stationary camera. The training normal data contains only pedestrians walking, while abnormal events are due to either the circulation of non-pedestrian entities (\eg car) or anomalous pedestrian motion patterns (\eg skateboarding). {(b) CUHK Avenue} includes 16 training and 21 testing videos, collected from a fixed scene. There are 47 abnormal events in total such as running, throwing bag, etc. {(c) ShanghaiTech} is very challenging that contains videos from 13 scenes with complex light conditions and camera angles. The overall number of frames for training and testing reach 274K and 42K, respectively. There exist 130 abnormal events in the test set, scattering in 17K frames.

\textbf{Evaluation criterion.} Following the popular evaluation metric in VAD literature~\cite{mahadevan2010anomaly,lu2013abnormal, liu2018future, gong2019memorizing,yu2020cloze}, we measure the Area Under the Receiver Operation Characteristic (AUROC) by varying the threshold over anomaly score. Higher AUROC indicates better VAD accuracy.

\textbf{Implementation details.}
We do not train our model on the whole video frames but the foreground objects. Following~\cite{yu2020cloze}, we extract all foreground objects for the training and testing videos. Each object is identified by an RoI bounding box. For each RoI, we build a spatial-temporal cube (STC) that contains not only the object in the current frame but also the contents in the same bounding box of previous $t$ frames, where $t=4$. The width and height of STCs are both resized to 32. Similarly, we extract the corresponding STCs for optical flows, which are estimated by FlowNet2.0~\cite{ilg2017flownet}. The extracted STCs for objects and optical flows are the buildingblocks used to train our model. During testing, the anomaly score of a frame is the maximum score of all objects in it. Considering the continuity of activity, the anomaly scores of a video are smoothed by a median filter whose window size is 17.

\begin{table}[t]
\renewcommand\tabcolsep{1.8pt} 
\centering
\caption{Comparison of frame-level anomaly detection performance with state-of-the-art methods. We calculate AUROC $\uparrow$ (\%) on UCSD Ped2 \cite{mahadevan2010anomaly}, CUHK Avenue \cite{lu2013abnormal} and ShanghaiTech \cite{luo2017revisit}. Numbers in bold indicate the best results.}
\vspace{0.3cm}
\begin{tabular}{c|cccc}
\hline
\textbf{Method}  & \textbf{\footnotesize UCSD Ped2}   & \textbf{\footnotesize  CUHK Avenue} & \textbf{SHTech} \\ \hline
Conv-AE~\cite{hasan2016learning}          & 90.0          & 70.2   & - \\
ConvLSTM-AE~\cite{luo2017remembering}     & 88.1          & 77.0   & - \\
GMFC-VAE~\cite{fan2020video}              & 92.2          & 83.4   & - \\
MemAE~\cite{gong2019memorizing}           & 94.1          & 83.3   &71.2 \\
MNAD-R~\cite{park2020learning}            & 90.2          & 82.8   &69.8 \\ 
\hline
Frame-Pred.~\cite{liu2018future}     & 95.4          & 85.1   & 72.8 \\
Conv-VRNN~\cite{lu2019future}        & 96.1          & 85.8   & - \\
MNAD-P~\cite{park2020learning}       & 97.0         & 88.5   & 70.5 \\
VEC~\cite{yu2020cloze}               & 97.3          & 90.2   & 74.8 \\ %
\hline
ST-AE~\cite{zhao2017spatio}               & 91.2          & 80.9   & - \\ AMC~\cite{nguyen2019anomaly}              & 96.2          & 86.9   & - \\
AnoPCN~\cite{ye2019anopcn}                & 96.8          & 86.2   & 73.6 \\

\hline
HF$^2$-VAD w/o FP                             & 98.8 & 86.8 & 73.1  \\ 
HF$^2$-VAD w/o FR                                  & 94.5          & 90.2 &76.0       \\
HF$^2$-VAD                           & \textbf{99.3} & \textbf{91.1} &\textbf{76.2}       \\ \hline
\end{tabular}
\vspace{-0.46cm}
\label{tbl:auc}
\end{table}

\begin{figure*}[ht] 
\vspace{-1.2cm}
\centering
 \subfigure[
 A Ped2 video with abnormal events: skateboarding and riding bicycle.]{\includegraphics[width=0.475\textwidth]{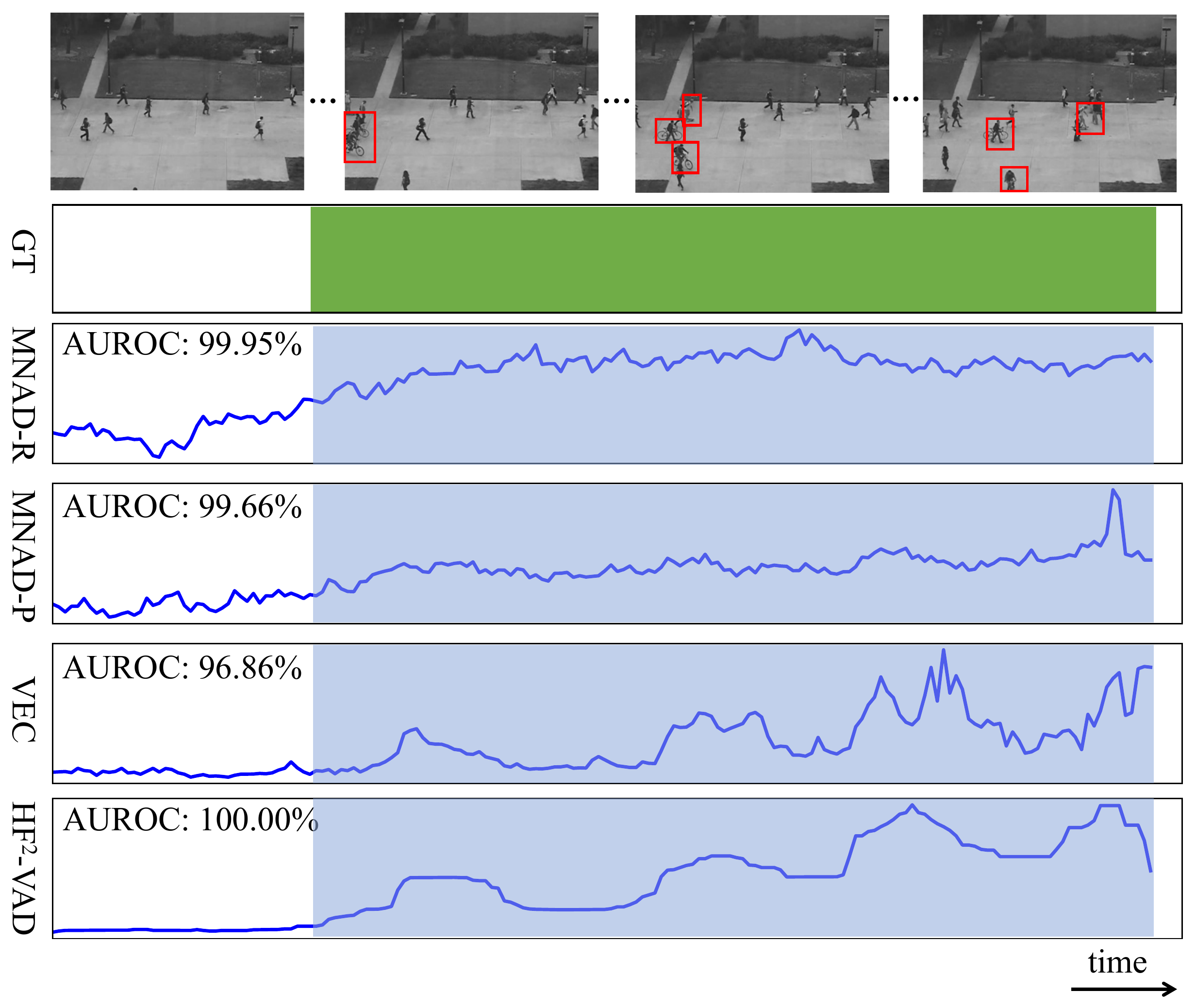}}
 \hspace{0.1cm}
 \subfigure[
 An Avenue video with the abnormal event: kid running.]{\includegraphics[width=0.475\textwidth]{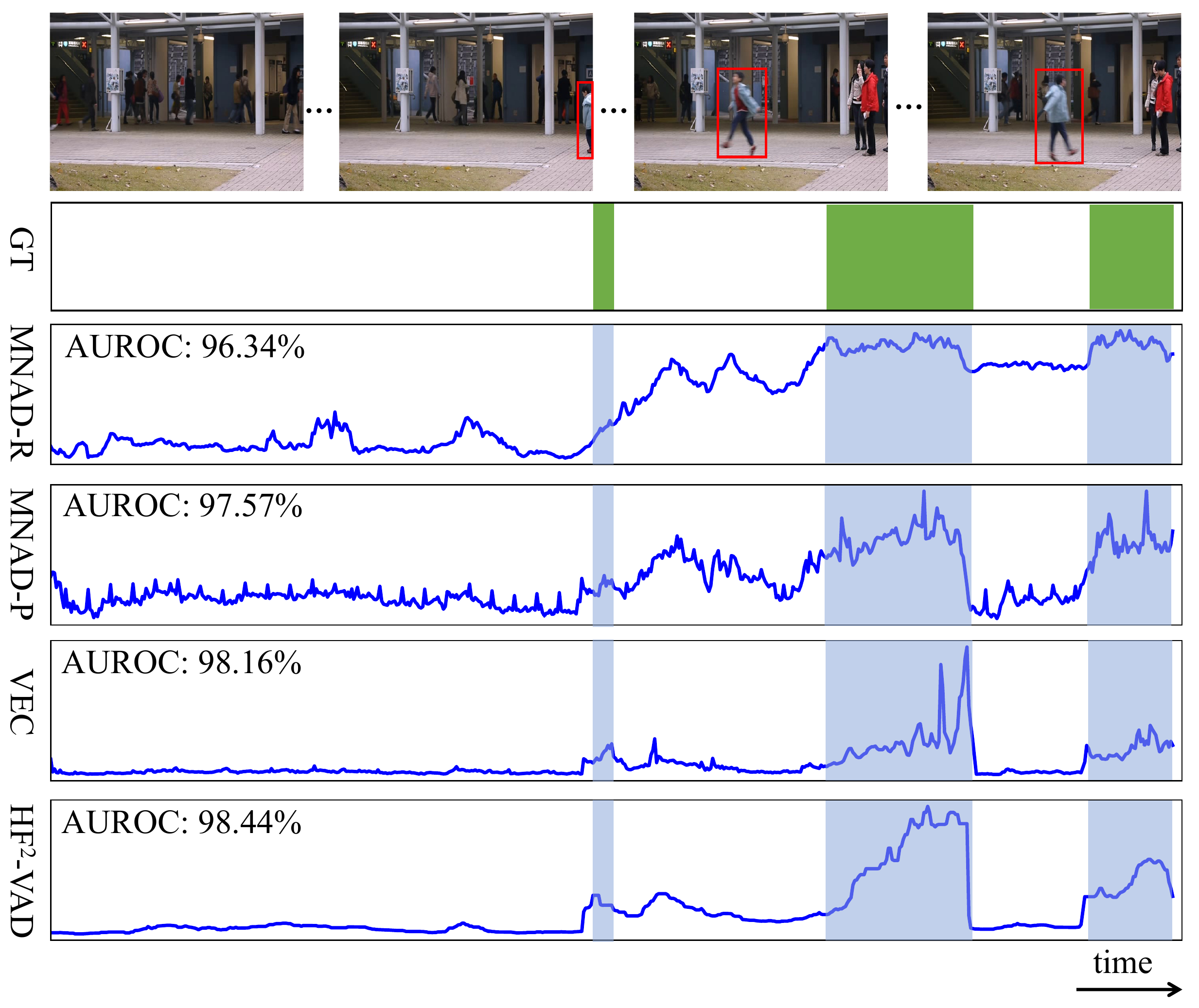}}\\
\caption{Two examples of anomaly detection comparison on USCD Ped2~\cite{mahadevan2010anomaly} and CUHK Avenue~\cite{lu2013abnormal}. From top to bottom, we show the sampled video frames, ground-truth abnormal sections (green regions are abnormal), result of MNAD-R~\cite{park2020learning}, result of MNAD-P~\cite{park2020learning}, result of VEC-VAD~\cite{yu2020cloze} and result of HF$^2$-VAD. Larger value in curves indicates more possible to be anomaly. Best viewed in color.} 
\label{fig:detecting_curves}
\vspace{-0.4cm}
\end{figure*}

We use PyTorch~\cite{paszke2017automatic} to implement HF$^2$-VAD and adopt Adam optimizer~\cite{kingma2014adam} with $\beta_1=0.9, \beta_2=0.999$ to optimize it. {The learning rate is initialized to $1e^{-4}$, decayed by 0.8 after every 50 epochs.} The slot number $N$ is set to 2K for all three datasets. The memory module numbers for Ped2, Avenue and ShanghaiTech are 3, 3 and 2, respectively. We train ML-MemAE-SC at first, then train CVAE model with the reconstructed flows, and finally finetune the whole framework. $\lambda_{recon}, \lambda_{ent}, \lambda_{CVAE}$ and $\lambda_{gd}$ are $1.0, 2e^{-4},1.0$ and $1.0$, respectively. The batch size and epoch number of Ped2, Avenue, and ShanghaiTech are set to (128, 80), (128, 80), and (256, 50), respectively. The error fusing weights $(w_r,w_p)$ for Ped2, Avenue and ShanghaiTech are set to (1.0, 0.1), (0.05, 1.0) and (0.02, 1.0), respectively.

\subsection{Results}
\textbf{Quantitative results.} We compare our proposed HF$^2$-VAD with state-of-the-art methods, including (1) reconstruction-based methods: Conv-AE~\cite{hasan2016learning}, ConvLSTM-AE~\cite{luo2017remembering}, GMFC-VAE~\cite{fan2020video}, MemAE~\cite{gong2019memorizing} and  MNAD-R~\cite{park2020learning}; (2) prediction-based methods: Frame-Pred.~\cite{liu2018future}, MNAD-P~\cite{park2020learning}, VEC~\cite{yu2020cloze} and Conv-VRNN~\cite{lu2019future}; and (3) hybrid methods including ST-AE~\cite{zhao2017spatio}, AMC~\cite{nguyen2019anomaly} and AnoPCN~\cite{ye2019anopcn}. In addition, we compare with two variants of our method, ``HF$^2$-VAD w/o FP" and ``HF$^2$-VAD w/o FR" which are formulated by removing frame prediction and flow reconstruction from HF$^2$-VAD, respectively. The results 
are summarized in Table \ref{tbl:auc}, and the performances of other methods are obtained from their original papers. 

As can be observed, the HF$^2$-VAD model achieves better results than state-of-the-art methods on all these three benchmarks, which demonstrates the effectiveness of our method. Especially, HF$^2$-VAD outperforms other hybrid methods by a large margin. For example, HF$^2$-VAD achieves the accuracy of 91.1\% on the CUHK Avenue dataset, but the best accuracy of previous hybrid methods is 86.9\% by AMC~\cite{nguyen2019anomaly}. Very interestingly, ``HF$^2$-VAD w/o FP" performs better on Ped2 than on CUHK-Avenue and ShanghaiTech. This is because optical flow provides more discriminative clues for the Ped2 dataset, which has also been observed by Yu \etal~\cite{yu2020cloze}. Oppositely, ``HF$^2$-VAD w/o FR" performs better on CUHK-Avenue and ShanghaiTech than on Ped2. Our full model combines the advantages of both, achieving the best results.

\begin{figure}[t]
\centering
\includegraphics[width=0.48\textwidth]{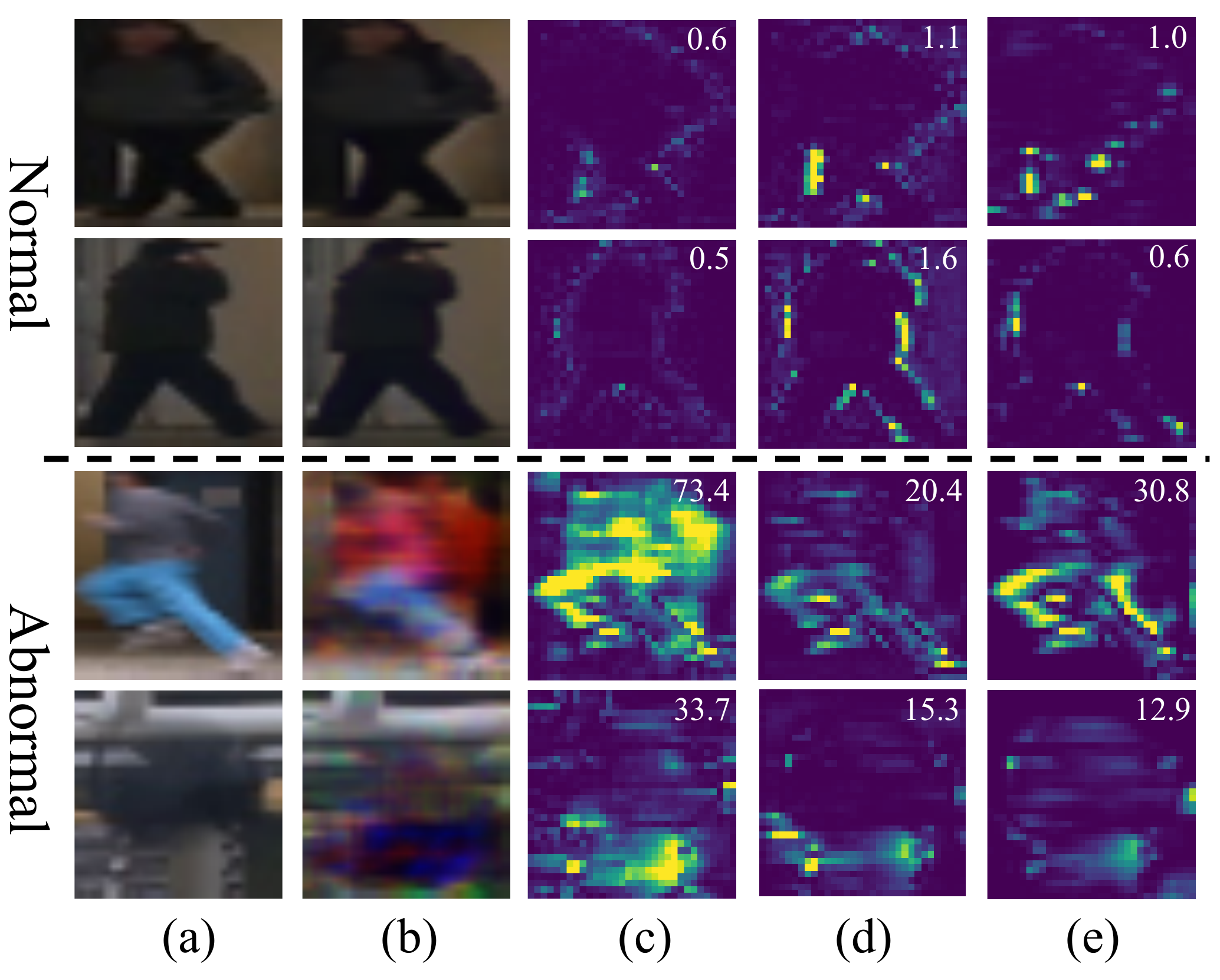}
\caption{Visualization examples of frame prediction comparison with different methods. From top to bottom, we show normal and abnormal data. From left to right, we show (a) the ground-truth, (b) prediction results of HF$^2$-VAD, (c) difference maps of HF$^2$-VAD, (d) difference maps of VEC~\cite{yu2020cloze}, and (e) difference maps of MNAD-P~\cite{park2020learning}. The numbers in each error map denote the corresponding sum-square-error between ground truth and the predicted frame. The lighter colors in error map denote larger prediction error. Best viewed in color.}
\label{fig:vis}
\vspace{-0.5cm}
\end{figure} 
\textbf{Qualitative results.}
Examples in Figure~\ref{fig:detecting_curves} show anomaly curves of two testing videos compared among MNAD-R~\cite{park2020learning}, MNAD-P~\cite{park2020learning}, VEC~\cite{yu2020cloze} and HF$^2$-VAD. An anomaly curve shows the anomaly scores of all frames of a video sequentially, by which we can more intuitively compare the performance of different methods. As can be seen, VEC and HF$^2$-VAD perform much better than MNAD-R and MNAD-P in normal sections, producing lower and more stable anomaly scores. HF$^2$-VAD is even better than VEC as it can better recognize abnormal events as shown in the abnormal durations where the anomaly scores computed by HF$^2$-VAD are higher than those by VEC. The AUROC values in these figures coincide with these intuitions.

In Figure~\ref{fig:vis}, we demonstrate several normal and abnormal images, showing the ground-truth in the first column and the predicted frames by HF$^2$-VAD in the second column. For saving space, we do not show predicted frames by VEC and MNAD-P, but instead show the difference maps between the ground-truth and the predicted frames by HF$^2$-VAD, VEC and MNAD-P in the last three columns respectively. The sum square errors of these difference maps are also shown. As we can see, HF$^2$-VAD produces less difference for normal images, while results in much larger differences for abnormal images. Taken the running person in the third row as an example, after the original abnormal optical flow being processed by ML-MemAE-SC, the flow fed into the CVAE is not consistent with the input images, yielding pixel shifting and color confusion as appeared in the predicted future frame.
\subsection{Discussion}\label{sec:experiment-ablation}
\begin{table}[t]
\vspace{-0.3cm}
\centering
\caption{Ablation study results on UCSD Ped2~\cite{mahadevan2010anomaly} dataset. The anomaly detection performance is reported in terms of AUROC $\uparrow$ (\%). Number in bold indicates the best result. }
\renewcommand\tabcolsep{1.5pt} 
\vspace{0.2cm}
\begin{tabular}{c|ccc|cc|c}
\hline
\textbf{} & \multicolumn{3}{c|}{\begin{tabular}[c]{@{}c@{}}Memory-augmented\\ Reconstruction Models\end{tabular}} & \multicolumn{2}{c|}{Prediction Models} & \multirow{1}{*}{ \vspace{-1cm} AUROC} \\ \cline{2-6} 
    & \begin{minipage}{0.11\columnwidth}
		\centering
		\includegraphics[width=\linewidth]{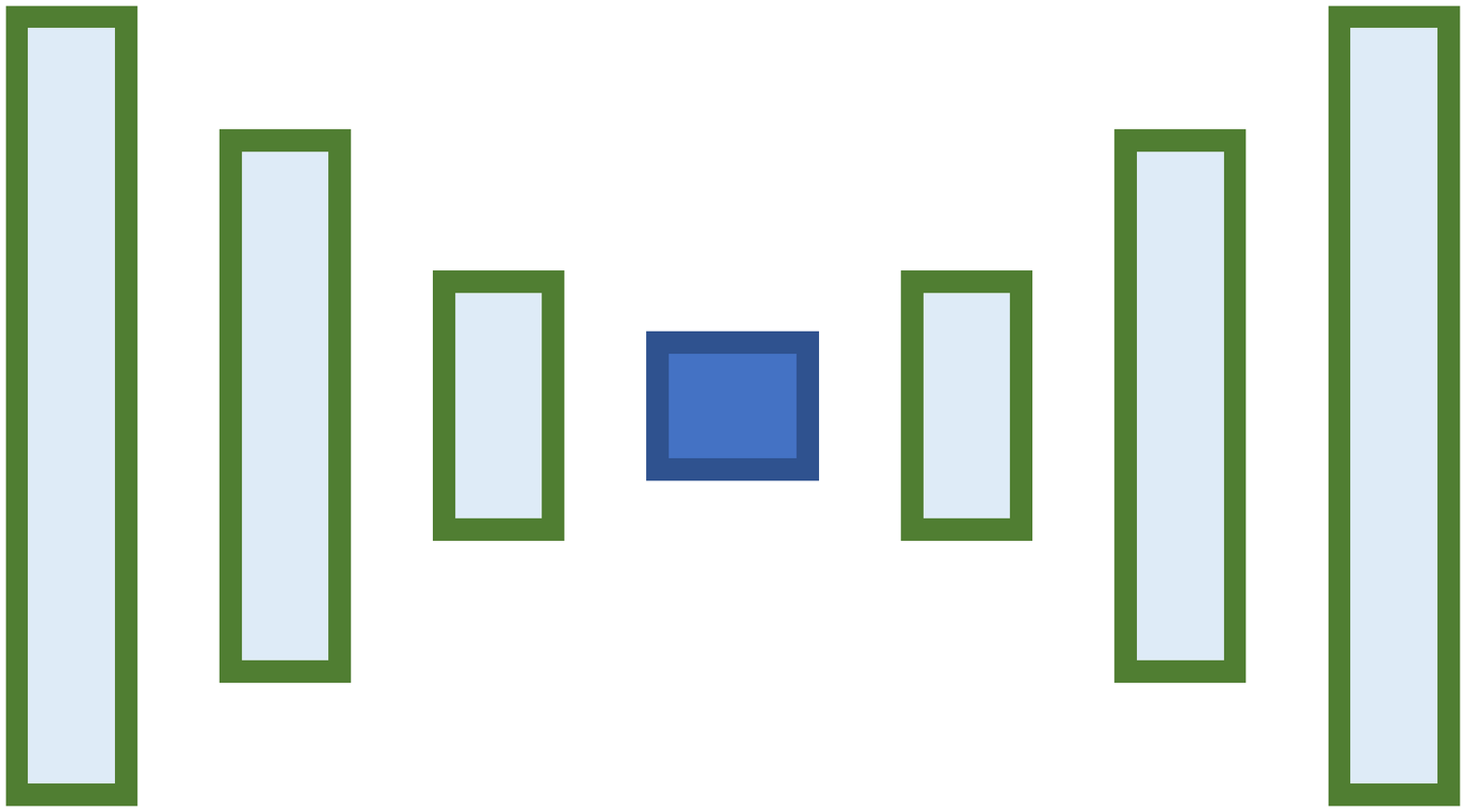}
	\end{minipage}
     & \begin{minipage}{0.11\columnwidth}
		\centering
		\includegraphics[width=\linewidth]{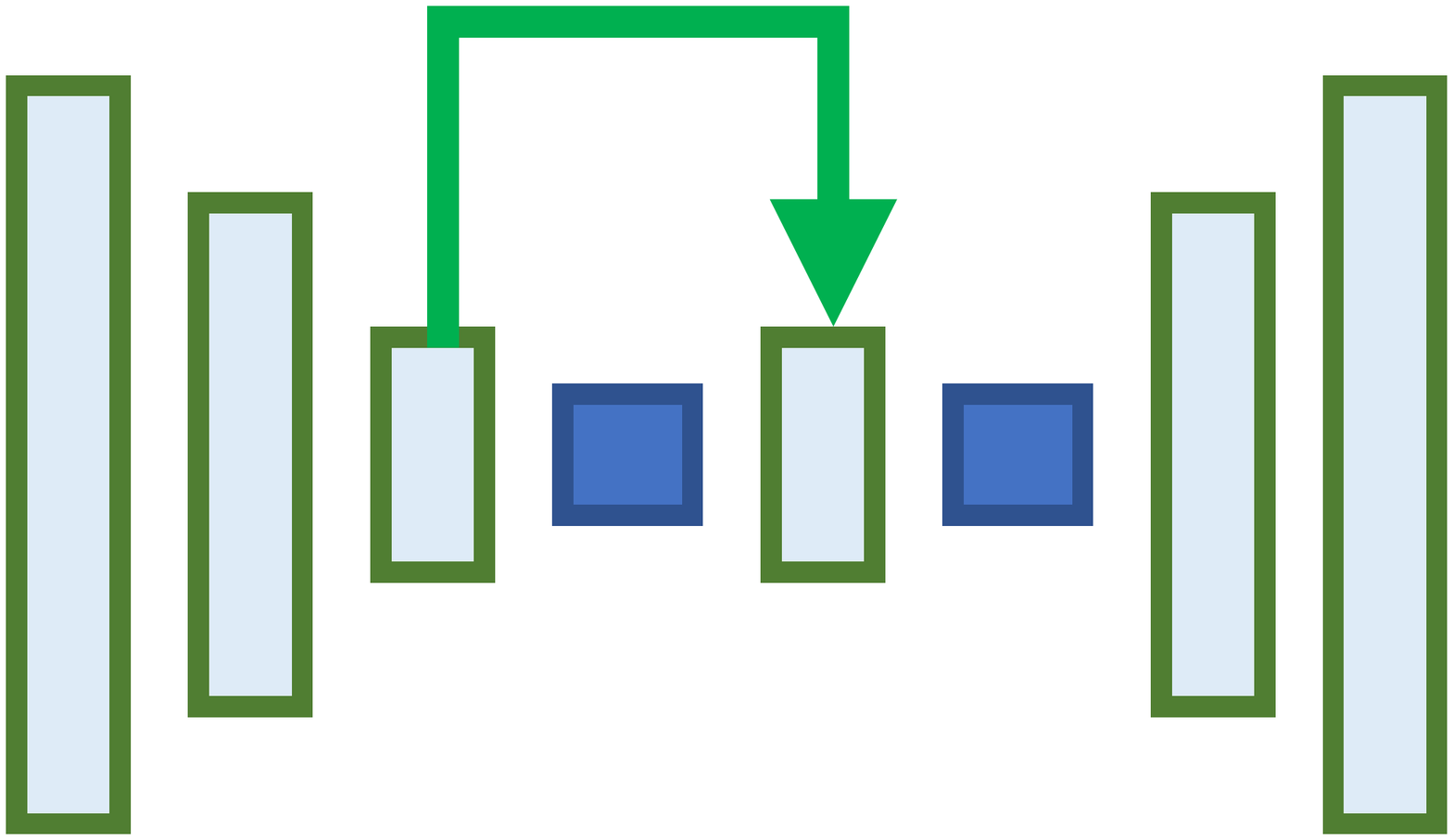}
	\end{minipage}
     & \begin{minipage}{0.12\columnwidth}
		\centering
		\includegraphics[width=\linewidth]{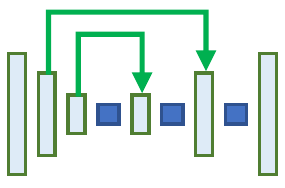}
	\end{minipage}              & \quad VAE    & CVAE     \\ \hline
	\multirow{3}{*}{Flow}     &\greencheck    &   &   &    &   & 96.27 \\
                                &   & \greencheck &   &      &  & 97.75 \\
                                &   &   & \greencheck &      &  & 98.81 \\ \hline
    \multirow{2}{*}{Frame}
    
  &   &   &   &\greencheck &     & 89.96 \\ 
  &   &   &   &  &\greencheck     & 94.48 \\ \hline
\multirow{3}{*}{Hybrid}         &\greencheck    &   &  &  &\greencheck    &96.91 \\
            &   &\greencheck  &   &   & \greencheck    &98.28 \\
            &   &   &\greencheck  &   &\greencheck   &\textbf{99.31} \\ \hline
\end{tabular}
\label{tbl:ablation}
\vspace{-0.35cm}
\end{table}
\textbf{Ablation investigation.}
To analyze the role of different components of HF$^2$-VAD, we conduct ablation study on UCSD Ped2 dataset~\cite{mahadevan2010anomaly} and report the anomaly detection performance in terms of AUROC in Table~\ref{tbl:ablation}. For the flow reconstruction part, we consider MemAE, ML-MemAE-SC with 2 memories and ML-MemAE-SC with 3 memories. For the prediction part, besides CVAE, we also investigate VAE for future frame prediction just based on previous frames. We then stitch the three reconstruction models with the CVAE to obtain three hybrid variants of our proposed method.

As can be seen from the table, just by reconstruction, the anomaly detection accuracy of ML-MemAE-SC with 3 memories is the highest, followed by ML-MemAE-SC with 2 memories, and finally MemAE, which are 98.81\%, 97.75\%, and 96.27\%, respectively. Just by prediction, CVAE outperforms VAE by a large margin from 89.96\% to 94.48\%, which shows that optical flow, as additional information, is critically necessary to increase the prediction accuracy. By integrating reconstruction and prediction, all the hybrid models are 
better than their corresponding reconstruction-only or prediction-only models, in which the hybrid method comprising ML-MemAE-SC with 3 memories and CVAE achieves the highest score among all the compared variants, which is 99.31\%.

\textbf{Computation time.} The experiments are performed on an NVIDIA RTX 3090 GPU and an Intel Core(TM) i9-7920X CPU @ 2.90GHz. As~\cite{yu2020cloze}, we need to preprocess the input video to extract all foreground objects and construct STCs for objects on the frame and optical flow. For our current implementation, the preprocessing phase averagely takes about 0.092s per frame. As for the anomaly detection phase, the model inference and anomaly score calculation together costs about 0.015s per frame. Overall, the running speed of HF$^2$-VAD is about 10fps.

\begin{figure}[t] 
\vspace{-0.3cm}
\centering
\includegraphics[width=0.475\textwidth]{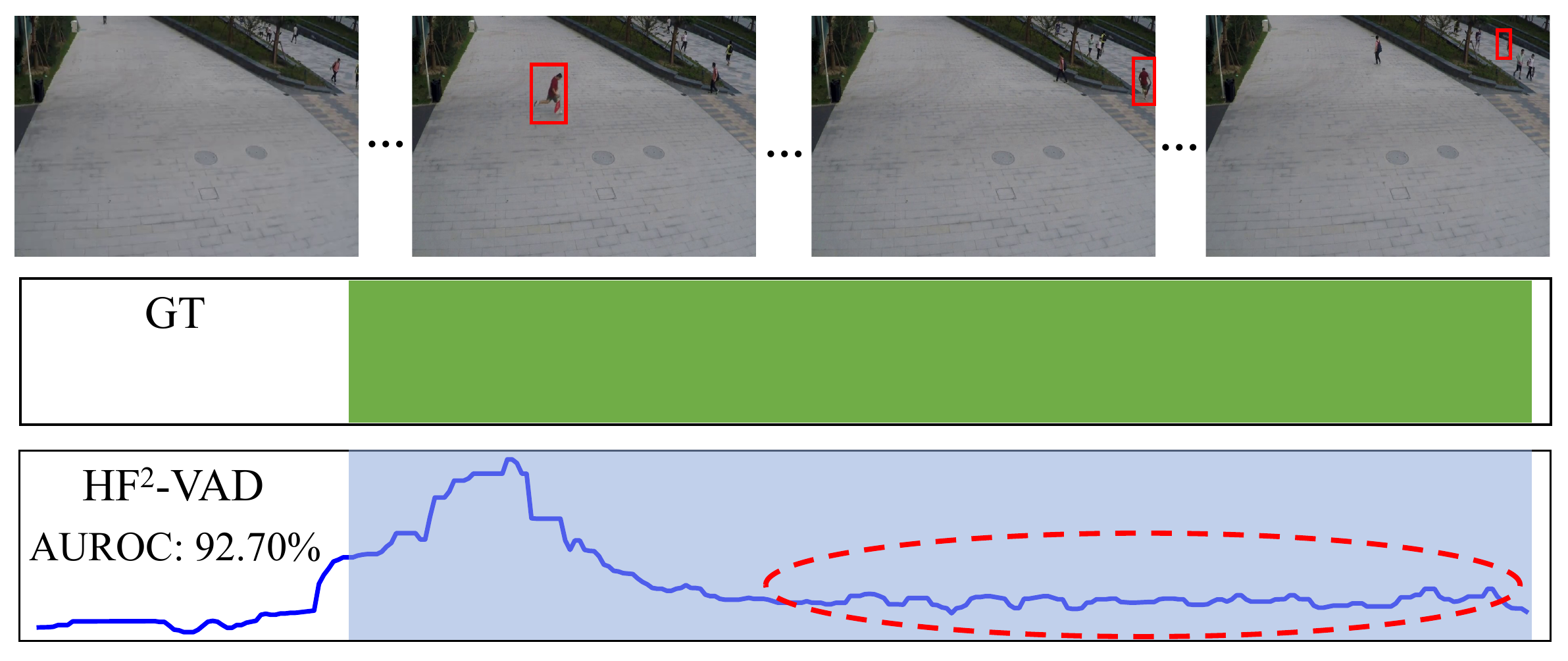}
\caption{A failure case of the proposed HF$^2$-VAD. The red dashed circle denotes the video section where the VAD performance is not good. Best viewed in color.} 
\vspace{-0.5cm}
\label{fig:failure_cases}
\end{figure}

\textbf{Failure case.} Figure~\ref{fig:failure_cases} shows a failure case of our method on a ShanghaiTech testing video in which a person is running. HF$^2$-VAD can easily detect it after the person comes into the view of the camera. But as the object runs farther away from the camera, the anomaly score becomes smaller and smaller. Our method fails on anomaly object that is far away, because a very far object, even though it is running, the absolute optical flow values are similar or even smaller than those of the normal objects near the camera. We conjecture that the scene depth is a very important variable that should be considered to further improve the VAD accuracy for the ShanghaiTech dataset, which would be investigated in future work.

\vspace{-0.2cm}
\section{Conclusion}
\vspace{-0.1cm}
In this paper, we have explored the possibility of combining reconstruction and prediction, which are the two most prevalent VAD paradigms nowadays, to obtain a hybrid VAD method for high-accuracy video anomaly detection. Experiments show that the proposed method outperforms previous reconstruction-only or prediction-only methods, and also is the most superior in the hybrid approach camp. Our integration strategy is novel, by utilizing a CVAE to predict future frame taking both previous video frames and optical flows as input, and proposing an effective reconstruction method, \ie the ML-MemAE-SC, to preprocess the flows in advance. That means our hybrid method is not a simple combination of reconstruction and prediction, but the reconstruction can effectively influence the prediction quality. It is the strong and inherent entanglement between the reconstruction and prediction that makes our method performs better than state-of-the-art approaches. 

\section*{Acknowledgement}
This research is sponsored in part by the National Natural Science Foundation of China (62072191, 61802453, 61972160), in part by the Natural Science Foundation of Guangdong Province (2019A1515010860, 2021A1515012301), and in part by the Fundamental Research Funds for the Central Universities (D2190670).

{\small
\bibliographystyle{ieee_fullname}
\bibliography{egbib}
}

\appendix
\noindent{\LARGE{\textbf{Appendices}}}
\section{Detailed Network Design}
\begin{figure*}[htp]
\centering
\includegraphics[width=1.0\textwidth]{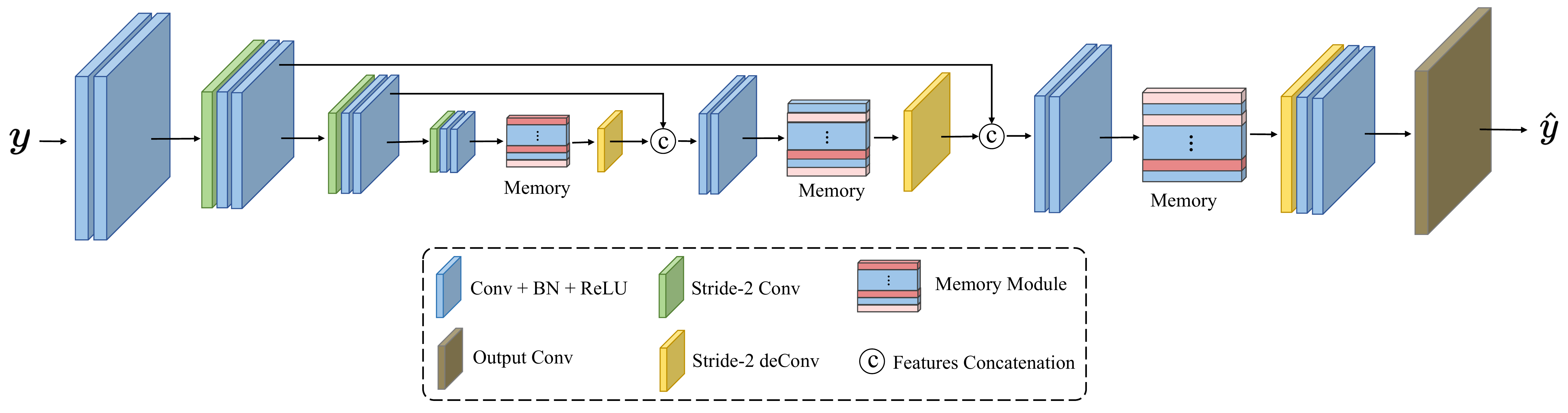}
\caption{Detailed network architecture of the ML-MemAE-SC for flow reconstruction.}
\label{fig:ml_memae_sc}
\end{figure*}

In Figure~\ref{fig:ml_memae_sc}, we illustrate the detailed network architecture of the ML-MemAE-SC for flow reconstruction. Each cube in the network is the output feature maps for the corresponding layer. ML-MemAE-SC contains 4 levels in total. The kernel size of all convolutional layers in the network is fixed to 3$\times$3. A basic convolution block contains a convolution layer, a batch-normalization layer and a ReLU activation layer sequentially.  The downsampling and upsampling layers are implemented by stride-2 convolution and stride-2 deconvolution, respectively.  The slot number of each memory module is fixed to 2K. Given an input flow with size (32,32,2), the feature maps sizes of each level are (32,32,32), (16,16,64), (8,8,128) and (4,4,256), respectively. 

\begin{figure*}[htp]
\centering
\includegraphics[width=1.0\textwidth]{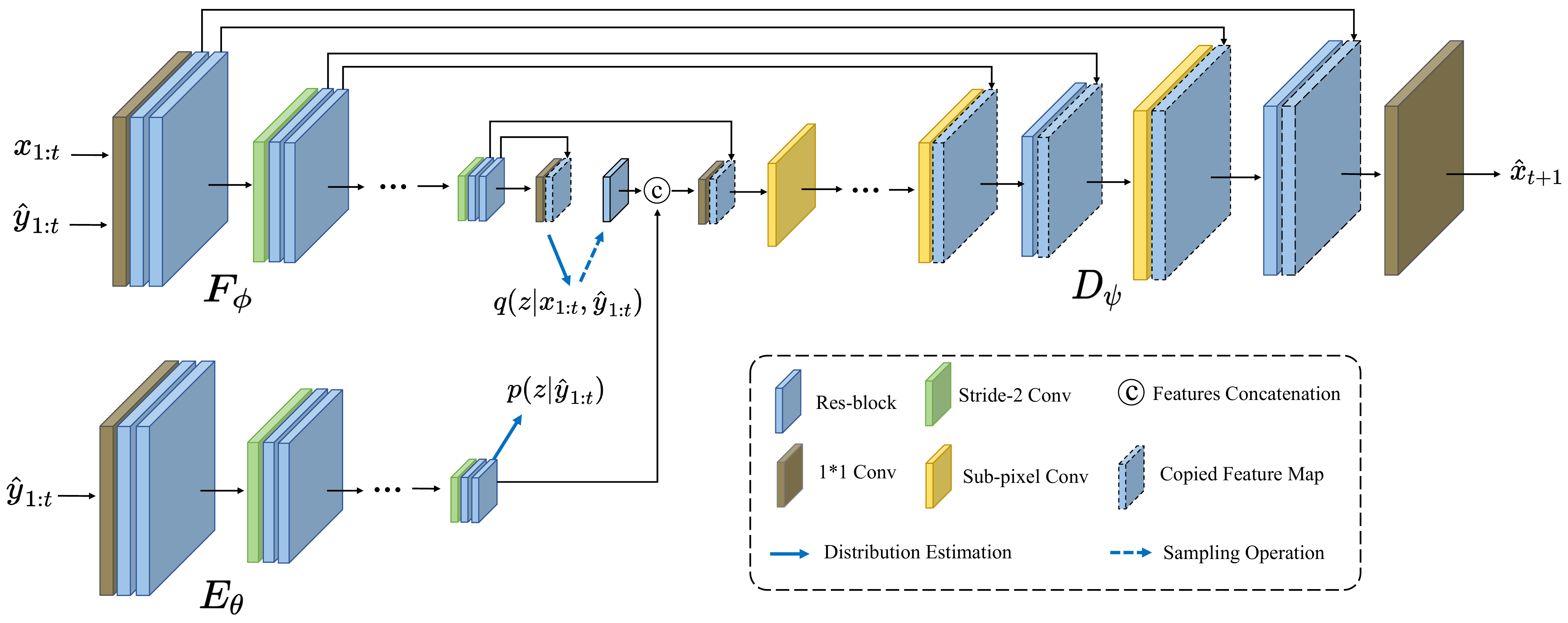}
\caption{Detailed network architecture of the CVAE for flow-guided future frame prediction.}
\label{fig:cvae}
\end{figure*}

In Figure~\ref{fig:cvae}, we illustrate the detailed network architecture of the CVAE for flow-guided future frame prediction. Each cube in the network is the output feature maps for the corresponding layer. As shown, we have two encoders $E_{\theta}$ and $F_{\phi}$ that share similar architecture, and one decoder $D_{\psi}$. Inspired by the Variational UNet proposed in~\cite{esser2018variational}, we add skip connections between $F_{\phi}$ and $D_{\psi}$ to help generating $x_{t+1}$. Following~\cite{esser2018variational}, the downsampling and upsampling layers are implemented by stride-2 convolution and subpixel convolution~\cite{shi2016real}, respectively. And each Res-block follows a similar setting as in~\cite{he2016identity}. Our CVAE model also contains 4 levels in total, and the corresponding feature map sizes of each level are (32,32,64), (16,16,128), (8,8,128) and (4,4,128), respectively. We concatenate the sampled $z$ with $E_{\theta}(\hat{y}_{1:t})$, which are sent to the decoder. Note that we utilize the last two bottleneck levels to estimate the distributions and sample data from them, and these two bottleneck levels share the same layer settings (please see the code for more details).

\section{Sampling strategies during test time}\label{appendix:sampling}
Conditional variational autoencoder (CVAE), as a generative model, can produce different output results when given different latent code during testing. We test two sampling strategies: (1) \textbf{stochastical way}, \ie sampling $z$ from the posterior distribution $q(z|x_{1:t},y_{1:t})$ randomly and (2) \textbf{deterministic way}, \ie using the mean of the posterior distribution $q(z|x_{1:t},y_{1:t})$ as the sampled $z$. For the UCSD Ped2~\cite{mahadevan2010anomaly} dataset, the AUROC of the latter strategy is $99.3078\%$, while the former way gives performance ranging from $99.3065\%$ to $99.3089\%$. This demonstrates that our model is robust though the predicted future frame is slightly different. But we still adopt the latter strategy to get statistically stable performance.

\section{Number of reconstructed flows to CVAE}
We have $t$ previous frames and $t$ corresponding optical flows ($t=4$ in our setting). We explore the performance of our method when inputting different number of reconstructed flows into the CVAE based prediction module. For example, we can input all $t$ reconstructed flows into CVAE, or just 1 reconstructed flow but $t-1$ original flows into CVAE. As shown in Table~\ref{tbl:flows}, there are totally 4 variants. The results show that our method with all the four reconstructed flows achieves the best VAD performance.

\begin{table}[]
\caption{Different number of reconstructed flows input into CVAE. As an example, $orig.\{1:t-1\} \quad recon.\{t\}$ means the flows from 1 to $t-1$ are original flows and the flow at time $t$ is reconstructed, which are fed into CVAE for the future frame prediction. Results are obtained on Ped2.}
\vspace{0.5cm}
\label{tbl:flows}
\small 
\setlength{\tabcolsep}{1mm}{
\begin{tabular}{c|c|c|c|c}
\hline
\textbf{} & \begin{tabular}[c]{@{}c@{}}orig.\{1:t-1\}\\ recon.\{t\}\end{tabular} & \begin{tabular}[c]{@{}c@{}}orig.\{1:t-2\}\\ recon.\{t-1:t\}\end{tabular} & \begin{tabular}[c]{@{}c@{}}orig.\{1:t-3\}\\ recon.\{t-2:t\}\end{tabular} & \begin{tabular}[c]{@{}c@{}}orig.\{1:t-4\}\\ recon.\{t-3:t\}\end{tabular} \\ \hline
AUROC   & 98.70\%        & 98.92\%       & 99.25\%                                                    & \textbf{99.31\%}      \\ \hline                                        
\end{tabular}
}
\end{table}

\section{Evaluation on UCF Crime}

The three VAD datasets evaluated in the paper consist of surveillance videos with static backgrounds, for which anomalies come from dynamic foreground objects. Therefore, we extract STCs and process each foreground object separately. But our method can also be applied to the entire video frames. To show this, we conduct experiment on UCF-Crime dataset~\cite{Sultani_2018_CVPR}. We select 10 videos for training and 6 for testing from UCF-Crime dataset. To be more specific, the training videos are \textit{Normal\_Videos165, 256, 267, 269, 279, 301, 355, 358, 489, 624}, and the test videos are \textit{Arson011, Explosion004, Explosion008, Explosion013, Explosion021, Shooting008}. We train the proposed HF$^2$-VAD model on the entire frames and the AUROC result is 83.50\% while that of VEC~\cite{yu2020cloze} is 81.12\%.

\section{Anomaly Detecting Cases }\label{sec:curves} 
We visualize more anomaly detection examples of the proposed HF$^2$-VAD framework, showing some anomaly curves in Figure \ref{fig:curves2} for UCSD Ped2~\cite{mahadevan2010anomaly}, CUHK Avenue~\cite{lu2013abnormal} and ShanghaiTech~\cite{luo2017revisit}. In each subfigure, the red boxes in video frames denote the ground truth abnormal objects, and we plot the anomaly score of each frame over time.  For a specific video, we calculate the AUROC under different model settings (higher AUROC means better anomaly detecting accuracy). We can observe that HF$^2$-VAD w/o FP or HF$^2$-VAD w/o FR can already detect most abnormal cases. Combining flow reconstruction and reconstructed-flow guided future frame prediction,  the HF$^2$-VAD performs even better,  producing relatively lower scores in the normal intervals and higher scores in the abnormal intervals.

\begin{figure*}[!ht] 
\centering
 \subfigure[Ped2 test video 01 with abnormal event: bicycle riding.]{
 \includegraphics[scale=0.35]{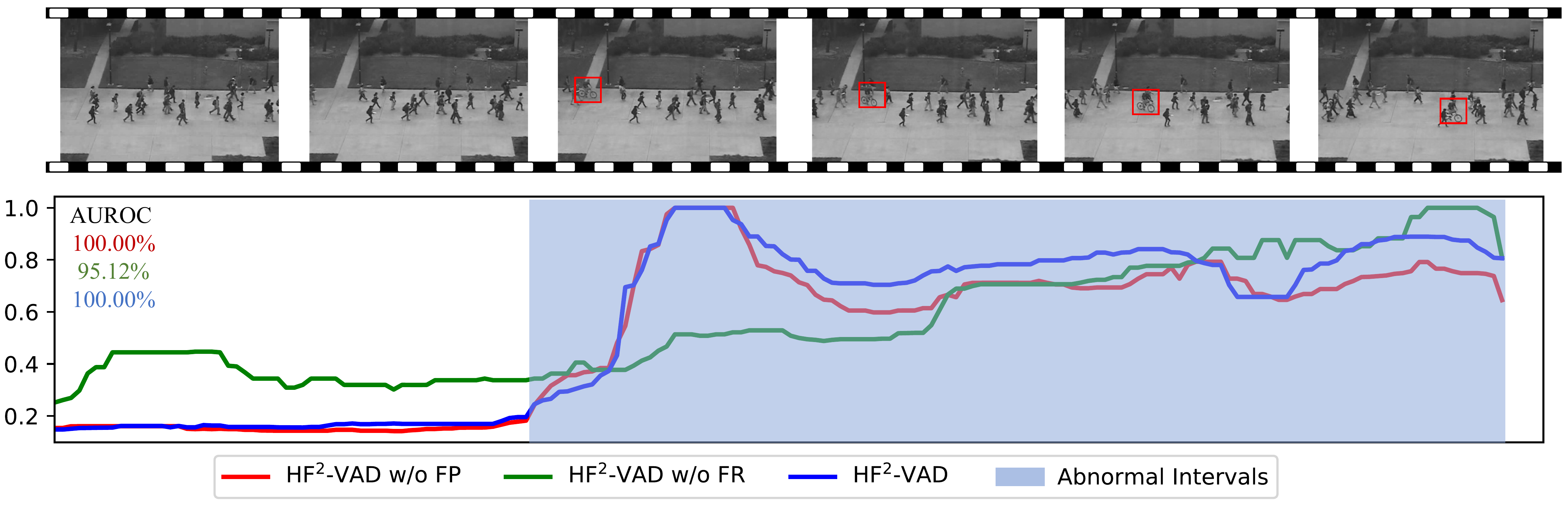}
 }
 \subfigure[Avenue test video 06 with abnormal events: wrong direction and throwing backpack.]{
 \includegraphics[scale=0.35]{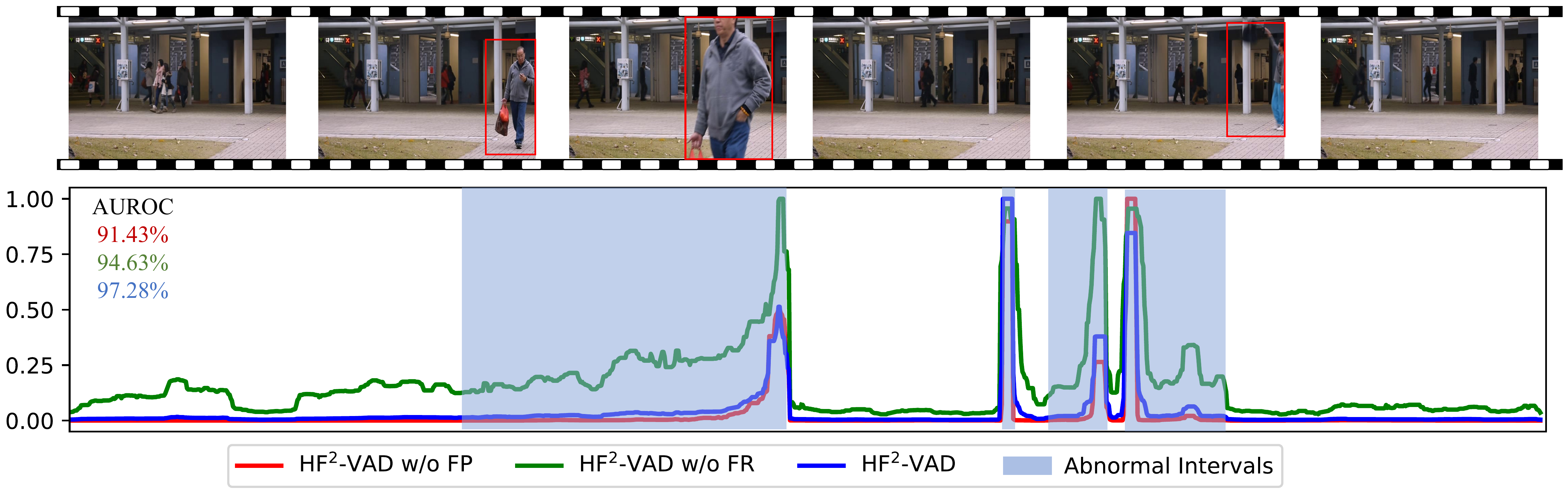}
 }
  \subfigure[Avenue test video 12 with abnormal event: throwing backpack.]{
 \includegraphics[scale=0.35]{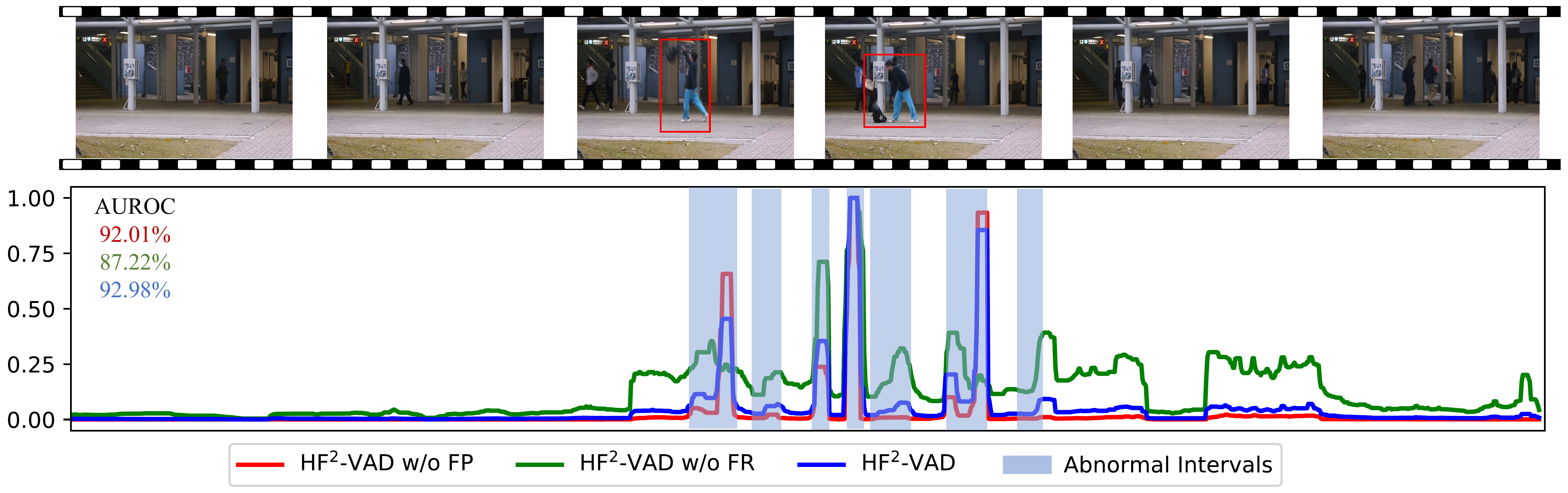}
 }
\end{figure*}
\addtocounter{figure}{-1}

\begin{figure*}[!ht] 
\addtocounter{figure}{1}
\centering
  \subfigure[ShanghaiTech test video 00\_0052 with abnormal event: bicycle riding.]{
 \includegraphics[scale=0.35]{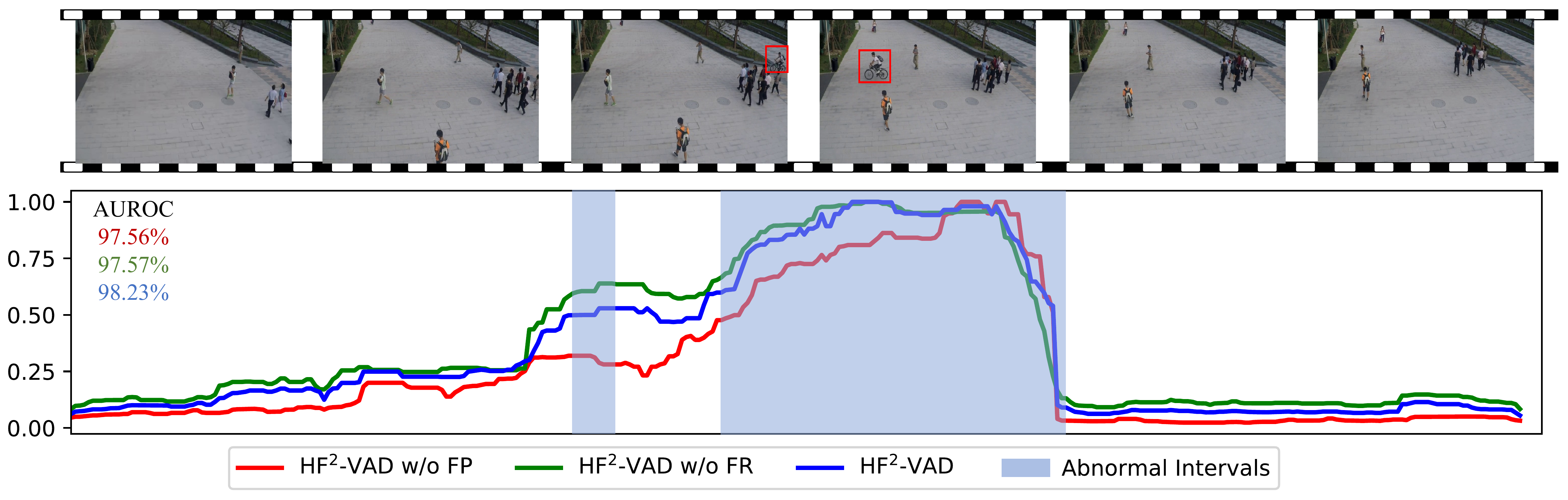}
 }
 \subfigure[ShanghaiTech test video 05\_0024 with abnormal event: fighting and chasing.]{
 \includegraphics[scale=0.35]{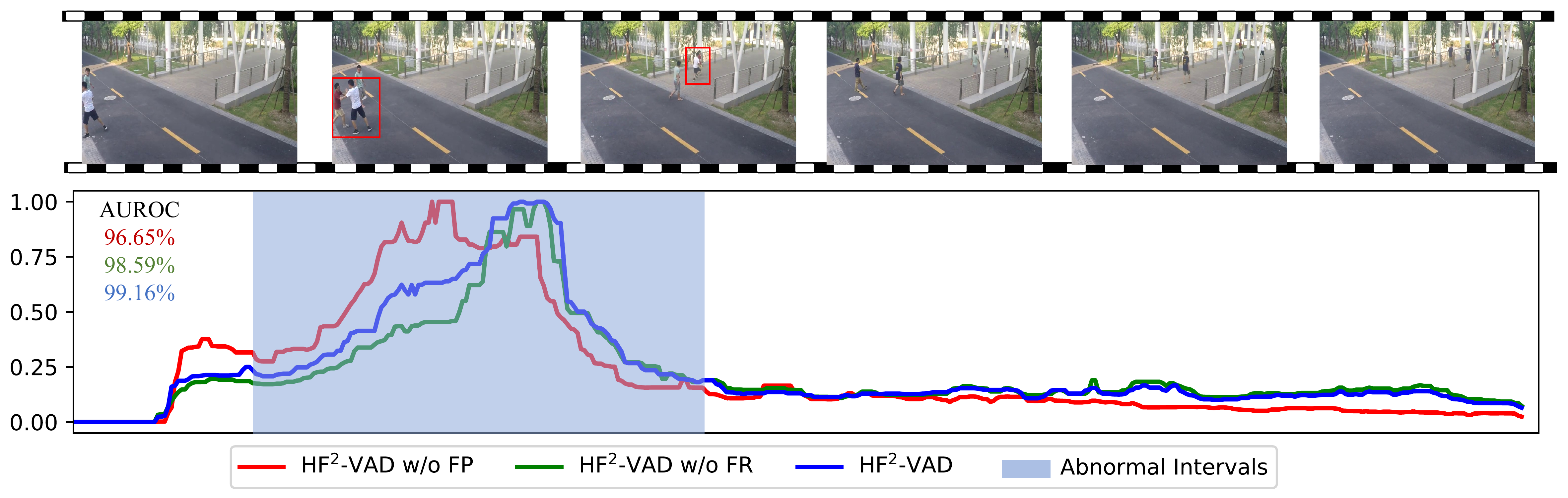}
 }
  \subfigure[ShanghaiTech test video 08\_0079 with abnormal event: running.]{
 \includegraphics[scale=0.35]{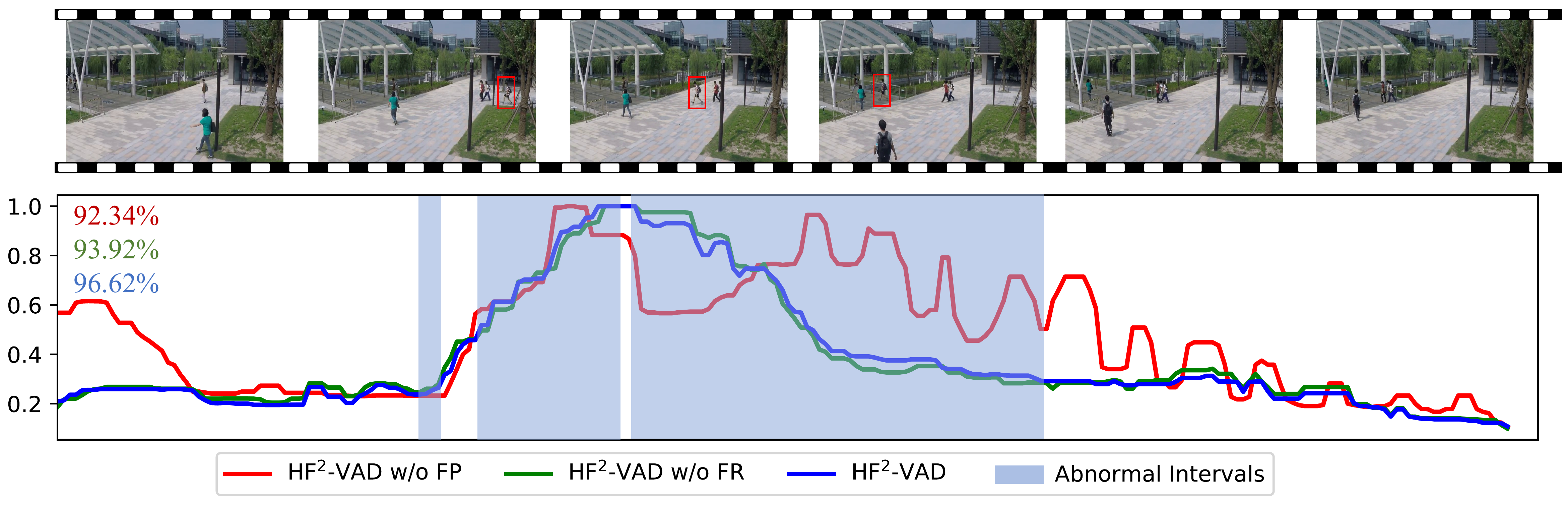}
 }\\
\caption{Anomaly detecting examples on USCD Ped2~\cite{mahadevan2010anomaly}, CUHK Avenue~\cite{lu2013abnormal} and ShanghaiTech~\cite{luo2017revisit}. The horizontal axis denotes time, while the vertical axis denotes anomaly score (higher value indicates more possible to be abnormal). The values in the upper left corner denote AUROCs under different model settings. Best viewed in color.} 
\label{fig:curves2}
\vspace{-0.4cm}

\end{figure*}

\section{More Qualitative Examples }\label{sec:mqe}
\begin{figure*}[t]
\centering
\includegraphics[width=1.0\textwidth]{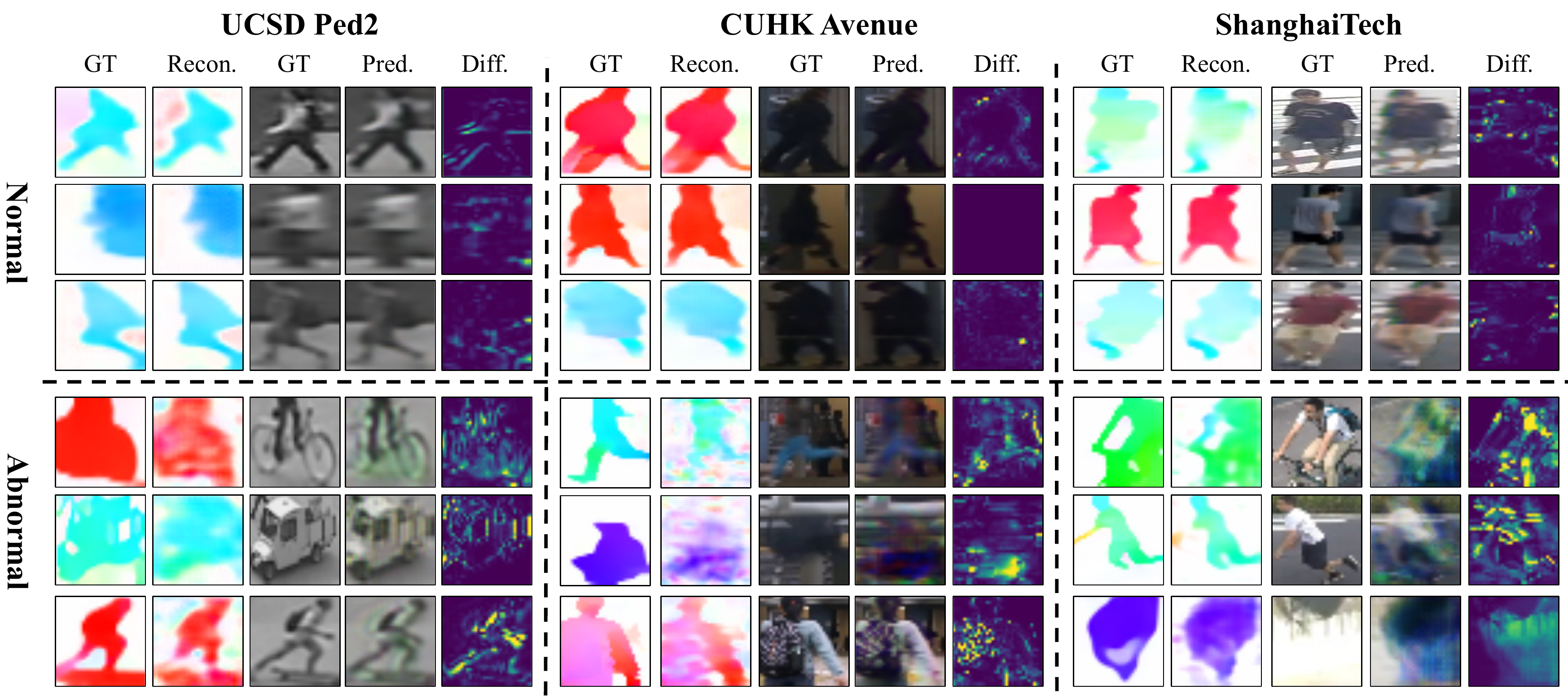}
\caption{Visualization of some flow reconstruction and future frame prediction examples on UCSD Ped2~\cite{mahadevan2010anomaly}, CUHK Avenue~\cite{lu2013abnormal} and ShanghaiTech~\cite{luo2017revisit} datasets. For each dataset, from left to right, we sequentially show the ground-truth flow, reconstructed flow, ground-truth future frame, predicted future frame and the prediction error map, respectively. The top and bottom regions show normal and abnormal samples respectively. The lighter color in the difference maps denotes larger prediction error. Best viewed in color.}
\label{fig:more_vis}
\end{figure*} 
We show more qualitative results of our proposed HF$^2$-VAD in Figure~\ref{fig:more_vis}, demonstrating some flow reconstruction examples and frame prediction examples. As can be seen, given a video event (\ie, flow spatial-temporal cube and frame spatial-temporal cube), the output of ML-MemAE-SC are inclined to be reconstructed as a combination of some normal motion patterns. We can clearly see that the normal flow patches are reconstructed well while the abnormal ones are not, which is an apparent clue to detect anomaly. Using the reconstructed motion as condition,  the predicted future frame for abnormal event is significantly different from the actual future, making it easier to be detected.

\end{document}